\documentclass[letterpaper, 10 pt, conference]{ieeeconf}  % Comment this line out if you need a4paper

\IEEEoverridecommandlockouts                              % This command is only needed if 
\usepackage{times} % assumes new font selection scheme installed
\usepackage{amsmath} % assumes amsmath package installed
\usepackage{amssymb}  % assumes amsmath package installed 
\usepackage{algorithm}
\usepackage[inkscapelatex=false]{svg}

\usepackage{amsthm}

\usepackage{appendix}
\usepackage{graphicx}
\usepackage{booktabs}
\usepackage{color}
\usepackage{booktabs} % 导入三线表需要的宏包

\usepackage{url}
\usepackage{multirow}
\usepackage{subfigure}
\usepackage{import}
\usepackage{transparent}
\usepackage[noend]{algpseudocode}
\usepackage{epsfig} % for postscript graphics files
\usepackage{xcolor}
\usepackage{pdfcolmk}
\usepackage{amstext}
\usepackage{ulem}
\usepackage{tikz}

\usepackage[linewidth=1pt]{mdframed}
\usepackage{soul} %删除线
\usepackage{graphicx} %插入图片的宏包
\usepackage{float} %设置图片浮动位置的宏包
\usepackage{subcaption}
\usepackage{caption}
\usepackage{multirow}
\usepackage{ulem}
\usepackage{makecell}
\theoremstyle{remark}
\newtheorem{remark}{Remark}

\title{\LARGE \bf
Whole-Body  Impedance Coordinative Control of  Wheel-Legged Robot on Uncertain Terrain
}

\author{
 Lei Shi$^{*}$, Xinghua Yu$^{*}$, Cheng Zhou$^{*{\dag}}$,  
Wanxin Jin,
Wanchao Chi, Shenghao Zhang,  Dongsheng Zhang,\\ Xiong Li$^{\dag}$, Zhengyou Zhang
\thanks{* denotes co-first author.
}
\thanks{\dag \  denotes corresponding author.
}
\thanks{
$^{1}$
The authors are with Tencent Robotics X, Shenzhen, Guangdong, China. {\tt\scriptsize \{chowchzhou,henricli\} @tencent.com}}%
\thanks{
$^{2}$
Wanxin Jin is with Arizona State University,
     Arizona, USA. {\tt\scriptsize wanxinjin@gmail.com}.  \ 
        }%
}
\begin{document}

\captionsetup{font={small}}

\maketitle
\thispagestyle{empty}
\pagestyle{empty}

%%%%%%%%%%%%%%%%%%%%%%%%%%%%%%%%%%%%%%%%%%%%%%%%%%%%%%%%%%%%%%%%%%%%%%%%%%%%%%%%
\begin{abstract}
% 这篇文章提出针对大惯性负载轮-腿式人形机器人的全身柔顺控制框架，使得机器人具有复杂地形的适应能力以及稳定的灵巧操作能力。 
This article propose a whole-body impedance coordinative control framework for a wheel-legged humanoid robot to achieve adaptability on complex terrains while maintaining robot upper body stability.
The framework contains a bi-level control strategy. The outer level is a variable damping impedance controller, which optimizes the damping parameters to ensure the stability of the upper body while holding an object.
% 首先给出包含机器人系统运动-动力学的等式约束，关节极限以及摩擦锥的不等式约束，基于轮足位置-力的地形信息更新，和基于无模型摩擦补偿方法的WBC框架。
%
%
The inner level employs Whole-Body Control (WBC) optimization that integrates real-time terrain estimation based on wheel-foot position and force data. It generates motor torques while accounting for dynamic constraints, joint limits, friction cones, real-time terrain updates, and a model-free friction compensation strategy.
%
%
% 紧接着我们给出了融合阻抗控制的全身柔顺控制方案，且为了维持上半身操作任务的稳定性，我们给出了基于变阻尼阻抗控制器，使得上半身可以有效的维持平稳。最后，我们采用最近刚研制的四足类人形机器人开展实验，实验结果证明，本文提出算法可以有效控制大惯量负载机器人，且在适应地形的同时有效的维持了上半身完成端水任务的稳定性。 
%
%
%
%
The proposed whole-body coordinative control method has been tested on a recently developed quadruped humanoid robot. The results demonstrate that the proposed algorithm effectively controls the robot, maintaining upper body stability to successfully complete a water-carrying task while adapting to varying terrains.
\end{abstract}

%%%%%%%%%%%%%%%%%%%%%%%%%%%%%%%%%%%%%%%%%%%%%%%%%%%%%%%%%%%%%%%%%%%%%%%%%%%%%%%%
% \keywords{ Wheel-legged Robots, Whole Body Control, Adaptive Impedance Control}

% \end{abst}

%%%%%%%%%%%%%%%%%%%%%%%%%%%%%%%%%%%%%%%%%%%%%%%%%%%%%%%%%%%%%%%%%%%%%%%%%%%%%%%%
\section{Introduction}

Humanoid robots serve as a crucial platform for research in embodied intelligence, combining dexterous manipulation, agile locomotion, and artificial intelligence, and they will play an increasingly important role in people's daily lives.
%
% Robots in daily life need to meet the requirements of stability, reliability and safety. 
Robots needs to adapt to uncertain terrain conditions, including varied ground materials like cobblestone and uphill or downhill slopes, while also maintaining passive stability in upper body tasks. This passive stability allows the robot to remain steady by passively responding to dynamic external forces, crucially enabling it to handle tasks involving delicate balance and object stability.

Achieving stability under unpredictable conditions is essential for robot applications in real-world environments, where the terrain and external forces may vary significantly. This adaptability is particularly critical for robots designed to navigate mixed terrain while carrying objects that require careful stability, such as trays with liquids or delicate equipment.
% At the same time, the robot body needs to have a certain load capacity.
%
%
% 目前较多的两足式机器人的的形态，在维持自身平衡需要单独设计复杂的控制算法，且负重能力有限，因而有必要对人居环境下机器人的形态重新定义。机器人的地形信息估计虽然可以接住外部传感器，如激光雷达，视觉等，但是其仍然无法给出机器人足部具体的接触点准确的坐标信息。而机器人足部-地面的接触信息对于全身控制来说至关重要。机器人在不确定地形下的移动和操作，需要保证力控制平稳定。这些研究方向目前都不够系统，因此有必要针对这些问题开展研究。 
At present, most of the bipedal  or two wheel humanoid robots need to design complex control algorithms to maintain their own balance, and their load-bearing capacity is limited. Therefore, it is necessary to redefine the robot's mechanical structure in the human-centered environment. 
%
% Besides, although the robot's terrain information estimation can derived from external sensors such as lidar and vision, it still cannot provide accurate frame information of the specific contact points of the robot's feet. 
% The contact information between the robot's feet and the ground is crucial for whole-body control. 
%
%
% The locomotion and manipulation of the robot in uncertain terrain requires stable force control. 
These research directions are not systematic enough at present, so it is necessary to conduct research on these issues.
%
%
% A central aspect of humanoid robotics is whole-body dynamics control, which underpins motion planning by enabling robots to adhere to dynamic constraints while achieving desired performance. 
%
% Whole-Body Control (WBC) provides a hierarchical framework for prioritizing tasks, enabling robots to allocate resources adaptively to critical operations while maintaining secondary functions.
%
% This approach enhances the robot's capacity for executing complex maneuvers and coordinated movements.
% %
% WBC is a powerful strategy for controlling robots in dynamic and unpredictable environments. By utilizing the robot’s full range of motion and adjusting task priorities in real-time, WBC enables the simultaneous execution of multiple tasks, ensuring both flexibility and adaptability. 

% 

\begin{figure}[t]
\captionsetup{font={footnotesize}}
\centering %
\includegraphics[width=0.45\textwidth]{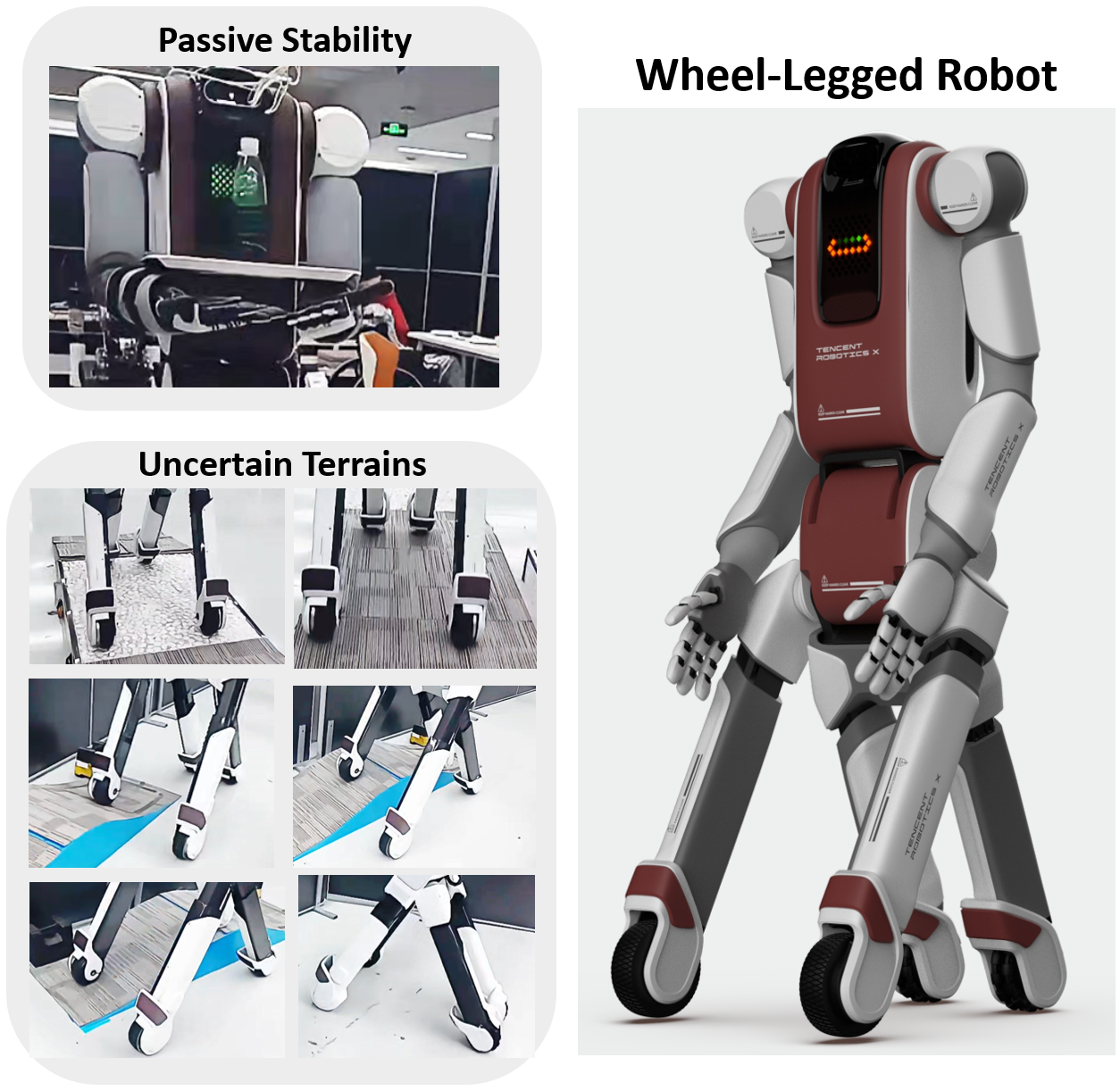}
% \caption{The locomotion-manipulation task of X-Man. We found that the humanoid robot has two impedance controllers in the same direction, such as $z$ direction, and an adaptive impedance control method is needed to maintain the stability of the manipulation task.
\caption{The newly developed wheel-legged robot, X-Man, with its 80 kg mass, introduces significant inertial challenges. Despite this, our work enables it to achieve upper-limb passive stability and demonstrate adaptability across varied uncertain terrains while maintaining balance.}
\vspace{-0.4cm}
\end{figure}

In this paper, we present a novel whole-body  control framework for our self-developed  robot, X-Man. 
%
% designed to integrate adaptive lower-body locomotion across varying terrains with upper-body manipulation. 
The main contributions are summarized as follows:
\begin{itemize}
% \item A novel control framework with the combination of whole body dynamics optimization control and impedance controller is proposed to solve whole body control of the large payload robot system;
% \item The online tire-terrain friction estimation and robot joint friction estimation is made, and they are compensated by the model-free medthod. 
% \item 
% Online detection of tire slippage, terrain height changes, etc. Based on this, the robot configuration is adjusted and drive wheel power distribution is 
% reallocated;
% \item Experiment validation on the real robot named as X-man, demonstrated the capability of whole body motion control, complex terrain adaptability.
% 在人形机器人的全身动力学控制器中融入了阻抗控制器，以适应不同地形以及操作任务。 并且为了保证上半身的操作精度，我们引入了变阻尼控制，以增强上半身操作物体时候的稳定性。 
\item The impedance coordinative controller is proposed to achieve passive stability and terrain adaption.
\item A novel whole-body control of a  humanoid robot  is proposed, including  whole-body dynamics optimization, terrain frame estimation and model-free  compensation.
\item Experiment validation on the  X-man, demonstrated the capability of whole body motion control, complex terrain adaptability.
\end{itemize}

The remainder of this paper is organized as follows: Section II reviews related work. 
Section III details the robot system.
The impedance coordinative controller and WBC controller are presented in Section IV, V, respectively. Section VI provides  experiments. Section VII concludes the paper.

\section{Related Work}
% \label{sec:II}
% 

% \textcolor{red}{A review of the forms and uses of robots.}

% \textcolor{red}{A review of the whole body control and gap， WBC.}

% \textcolor{red}{A review of the  WBC + impedance.}

\subsection{The Humanoid Robot}

Currently, there are three main types of humanoid robots, namely bipedal, two-wheel  type, and quadrupedal~\cite{tong2024advancements}.
Typical representatives of bipedal humanoid robots include Altas, TORO~\cite{kuindersma2016optimization,mesesan2019dynamic}. 
%
% Its main research purpose is used in complex scenes, such as rugged outdoor roads.
%
Even the simplest movement of standing up requires a lot of energy for a bipedal robot.
For two-wheeled humanoid robots, such as $\text{EVE}^{\text{R3}}$~\cite{gupta20192d}. This type of robot  is mainly used on regular roads.
For quadruped humanoid robots, such as Centauro and GITAI~\cite{kashiri2019centauro,ma2023advances}. This type of robot is mainly used in complex scenes, but due to its large weight, the overall dynamic performance and environmental adaptability are poor.
%
% 就用途而言，目前的人形机器人主要还是用于户外，甚至空间环境探索中。对于居家场景中的机器人形态探索目前研究较少。
In terms of usage, current humanoid robots are mainly used in outdoor and even space environment exploration~\cite{diftler2011robonaut}.

% 综上所示，目前大多数人形机器人在用途上依然有限，但长远来看，它们将会在家庭、医院等场景中承担越来越多的日常任务。此类场景下的机器人形态也需要重新定义。
In summary, there is currently little research on the exploration of robot morphology in human-centered environment.  The robot form in such scenarios  needs to be redefined.

%
% Besides, the whole-body dynamics framework was also applicable to purely manipulative tasks~\cite{dehio2022enabling},  quadrupedal robot's dynamic locomotion~\cite{bellicoso2017dynamic}, planetary rover system~\cite{bussmann2018whole}, and humanoid robots with parallel chains~\cite{sovukluk2023whole}.
% In Ref~\cite{ramuzat2021comparison}, the position and torque based whole body controller was compared on the platform of humanoid robot TALOS.

\subsection{The Whole Body Control}

% WBC可以控制多自由度机器人，并且可以让机器人在非确定环境下，安全，柔顺和稳健地抵抗未知干扰。
% WBC can control multi-DOF robots and enable them to resist unknown interference safely, compliantly and robustly in uncertain environments~\cite{wensing2023optimization}.
%
%
% Research on whole-body control is mainly divided into kinematics-based whole-body control and dynamics-based whole-body control.
The  whole-body controller can be divided into least-square(LS) based solver and Quadratic Programming(QP) controller based solver~\cite{wensing2023optimization}.
As for the LS solution,
the inverse dynamics based WBC of floating base robot with contact constraints was  studied~\cite{righetti2011inverse}.
The divergent component of motion(also known as instantaneous capture point) and passive based whole body controller is utilized to achieve dynamic walking and even dynamic multi-contact transitions tasks for TORO~\cite{mesesan2019dynamic}.
And the QP-based optimization frame can be divided into two types as weighted quadratic programming(WQP) and hierarchical quadratic programming(HQP)~\cite{hutter2014quadrupedal}.
The hierarchical optimization is also utilized to achieve perception-less terrain adaptation for ANYmal~\cite{bellicoso2016perception}.
WBC is also utilizded to tracking the commanded COM (Center of mass) trajectories generated via 3D SLIP(spring loaded inverted pendulum)~\cite{sovukluk2023highly}.

Besides, for robots with large inertia, the nonlinear characteristics of the robot joints are obvious~\cite{ verbert2015adaptive}.
%
% Model-based approaches can be unstable during the robot's dynamic-static friction switching process.
%
% And the static friction models include: Coulomb model, static friction and Coulomb friction models, stribeck model, karnopp model, Armstrong model. 
%
% The dynamic friction models mainly include Dahl model and LuGre model.
%
%
% As for the model free method, the PD method is studied to avoid stick slip~\cite{dupont1994avoiding, verbert2015adaptive}.
% ~\cite{dupont1994avoiding, verbert2015adaptive,kim2019model,le2008friction,kim2015disturbance}.
%
%
%
% We can see, the current humanoid robot WBC does not take friction and terrain information into consideration. This paper is one of the few works that have applied WBC to large inertia  robots.
% 目前对大惯量人形机器人的全身动力学控制并没有很系统性的研究，尤其是大惯量带来的非线性项，地形不确定和大惯量带来的力控制扰动问题。
At present, there is no systematic research on the WBC of large-inertia humanoid robots.
% , especially the nonlinear terms, terrain uncertainty and force control disturbance problems.

%
%

% %
%
%
% 机器人的笛卡尔阻抗控制也有从多优先级的角度来研究，其中，有基于雅可比伪逆的方案，也有通过QP求解器解算的方案。

\subsection{The Impedance Control}
The combination of impedance control and whole body control can achieve  the whole body complaince.
The Cartesian impedance control of robots has also been studied from a multi-priority perspective, including solutions based on Jacobi pseudo-inverse~\cite{dietrich2019hierarchical} and solutions solved by QP solver~\cite{hoffman2018multi}.
The impedance control and the optimization-based whole-body control framework is implemented to achieve stable biped walking~\cite{jo2022robust}.
    And the optimal impedance planning by given the Cartesian inertia and computes the stiffness and damping gains without relying on force/torque sensor is also studied~\cite{pollayil2022choosing}. 
    % 然而人形机器人全身自由较多，在同一方向上存在多个任务，如高度任务和双臂的操作竖直方向等，现有的优先级阻抗控制无法处理稳定性问题。 而本文则受到悬浮背包原理的启发，提出基于变阻尼阻抗控制结合WBC的方法来实现人形机器人地形适应的同时，双臂的操作稳定性。

    As for humanoid robots, there will be two impedance controllers in the same direction. Force control in this impedance-coupled situation in humanoid robots has rarely been studied before.

\begin{figure}[t]
\centering
\subfigure[Robot tasks.]{
\begin{minipage}[t]{0.5\linewidth}
\centering
\includegraphics[width=2.0in]{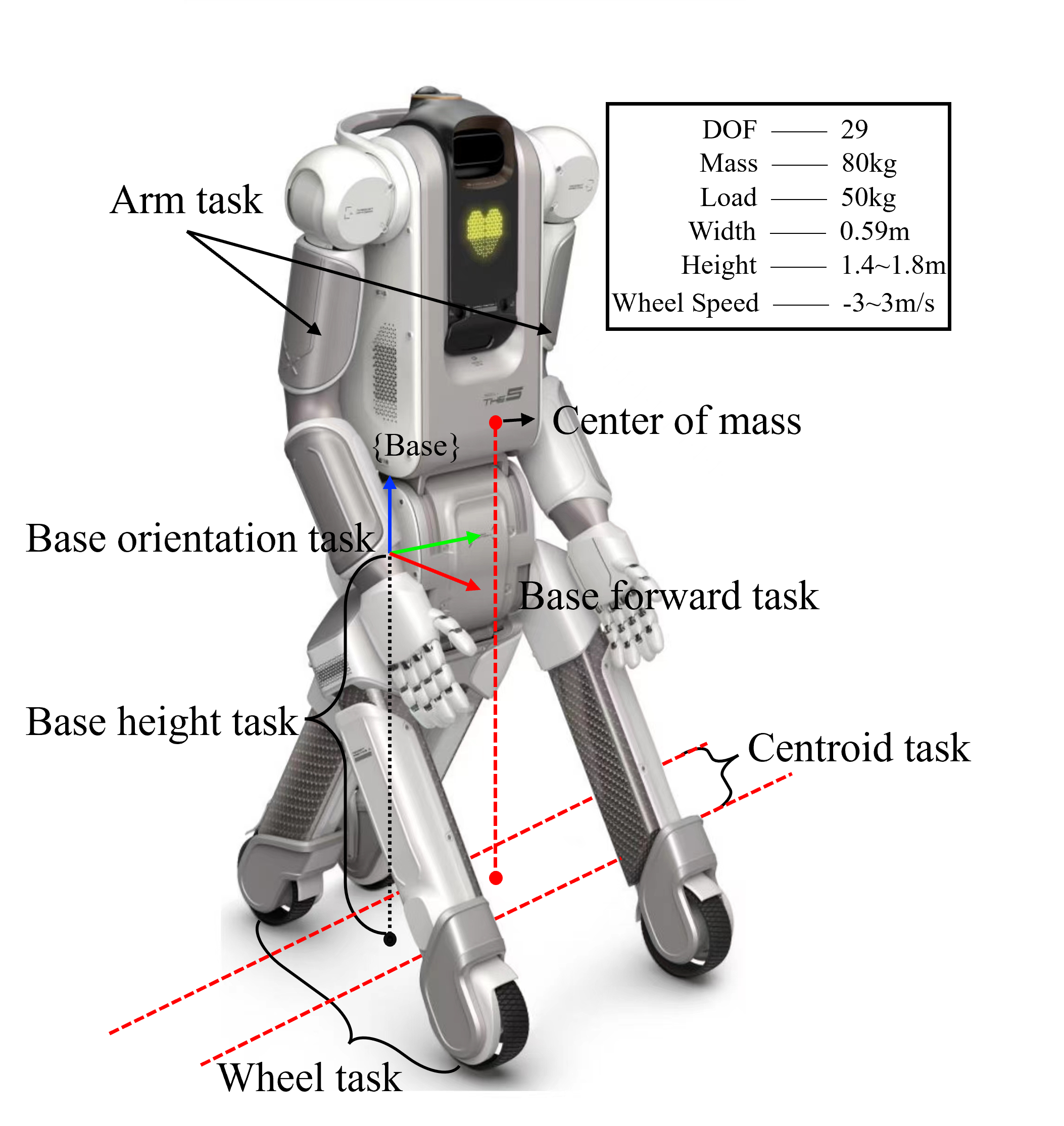}
\end{minipage}%
}%
\subfigure[Robot joints.]{
\begin{minipage}[t]{0.5\linewidth}
\centering
\includegraphics[width=1.4in]{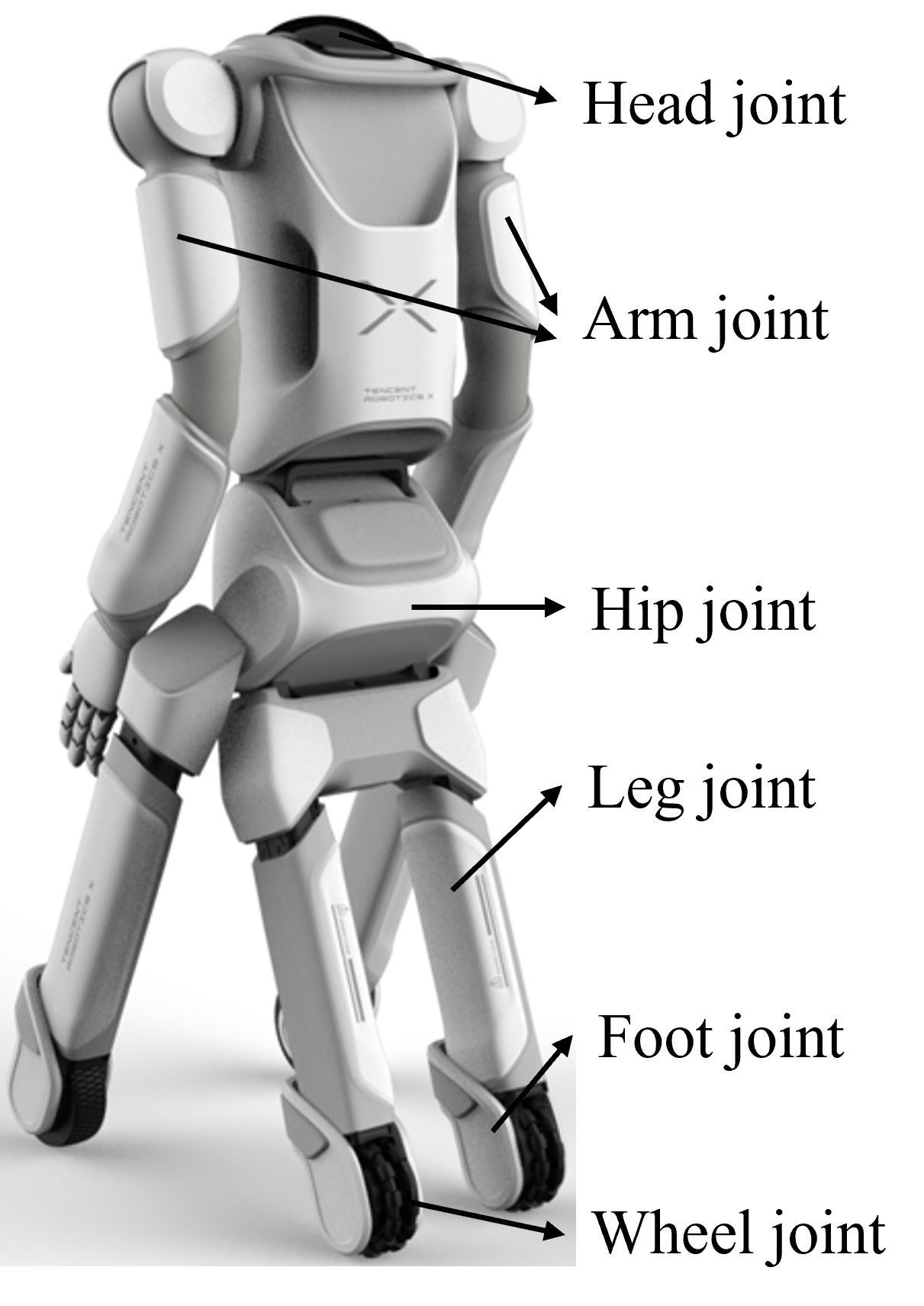}
\end{minipage}%
}%
\centering
\captionsetup{font={footnotesize}}
\caption{Schematic diagram of X-Man. Robot control tasks including: base position task(height, forward), base orientation task, wheel-Centroid task, Centroid task, arm task. In addition, we define the head orientation as the front, and according to the configuration characteristics of the robot, it can be divided into front legs and hind legs, front wheels and rear wheels, front hip and rear hip. The robot's legs are mobile joints, and other parts are rotational joints.}
\vspace{-0.5cm}
\end{figure}

% \begin{figure}[t]
% \begin{minipage}[t]{0.5\textwidth}
%      \centering
%     \includegraphics[width=0.5\linewidth]{figure/wbc_tasks.tif}
%     \caption{Robot tasks.}
%     \label{fig:enter-label}
% \end{minipage}
% \hfill
% \begin{minipage}[t]
%     \centering
%     \includegraphics[width=0.5\linewidth]{figure/robot_joints.tif}
%     \caption{Robot joints.}
%     \label{fig:enter-label}
% \end{minipage}
% \end{figure}

% \begin{figure}[t]
%   \centering
%   \captionsetup{font={footnotesize}}
%   \def\svgwidth{\columnwidth}
%   \includesvg[width=0.75\linewidth]{wbctasks.svg}
%   \caption{Schematic diagram of X-Man. Our wheel-legged robot consists of four-wheel, four-leg, and two-hip, wait-pitch, waist-yaw, and two-arm with 7-DOF respectively.
%   %
%   Among of these, the robot's legs are mobile joints, and other parts are rotational joints.
%   %
%   %
%   The two outer wheels of X-Man are active wheels, driven by hub motors; the two inner wheels are passive wheels, driven by universal wheels, allowing the robot to move freely in the $x$ and $y$ directions of its body frame.
%   %
%   %
%   Besides,
%   %
%   control tasks including: base position task, base orientation task, wheel-Centroid task, Centroid task, upper-body task.}
% \end{figure}

% \begin{figure}[t]
% \captionsetup{font={footnotesize}}
% \centering
% \includegraphics[width=0.75\linewidth]{figure/m1.png}
%         \caption{Schematic diagram of X-Man. It is mainly composed of wheels, legs, hips, waist, arms, hands and head.}
% \label{fig:wbc-tasks}
% \end{figure}

\section{Hardware and System Overview of X-Man}
% \section{Control Objective and Control Strategy}
\label{sec:III}

\subsection{Hardware Design}
This paper aims to design a humanoid quadruped robot for human-centered environment.
Common bipedal humanoid robots require a lot of energy and the design of complex control algorithms to achieve robot locomotion, and the system has poor robustness and weak anti-disturbance capabilities.
In a human-centered environment, the  robot need to gurantee terrain adaptability, and stable upper body manipulation capabilities.
The robot adopts a wheel-leg composite design, which can greatly expand the range of movement and enhance movement stability, while still retaining the characteristics of a legged robot.
The robot's lower body joints, including the waist, hip, leg, and foot joints, are equipped with joint torque sensors. The upper body arms are covered with tactile skin.
In addition, we define the head orientation as the front, and according to the configuration characteristics of the robot, it can be divided into front legs and hind legs, front wheels and rear wheels, front hip and rear hip.

\subsection{Control System Overview}

\subsubsection{Control Objective}
% talk about objective of the control system: stablize the upper body and minmize the energy of the lower-body.
We derive the X-man controller with control objectives divided into two main components: (A) upper-body control, which includes passively stabilizing the upper limbs and minimizing the energy consumption of the lower limbs due to body oscillation; and (B) whole-body control(WBC), which involves tracking a desired trajectory. 

For (A), the energy expenditure resulting from the force provided by the lower body to the upper body is computed by integrating the mechanical power \(\rho_1(t)\) over a full cycle \(T\). To maintain the stability of the upper limbs during holding an object, a stability term \(\rho_2(t)\) is introduced. Thus, the total mechanical work associated with upper limb is expressed as:
\begin{equation}
E_k = \int_0^T \rho_k(t) \, dt. \ \ \ k=1,2
\end{equation}
The decision variable for upper limb is the variable damping parameter.

For  (B), whole-body control aims to achieve accurate trajectory tracking by applying joint torques. The objective function for WBC consists of a state regularization term combined with a body vibration suppression term to ensure stability and precise movement throughout the trajectory.

% (Lei this line: to generally write E_1 E_2 expression)
% mention about the decision variable:damping parameter+ torque, specific to our xman. 

\subsubsection{Our Control Strategy}
% in order to solve the above control problem, we choose to decouple the control into two subproblem based on our platform. First, sub control problem: we only optimize with the damping parameter. Second: sub probelm:based on the optimized damping, we generate the control torch.

\begin{figure}[t]
\captionsetup{font={footnotesize}}
\centering %
\includegraphics[width=0.48\textwidth]{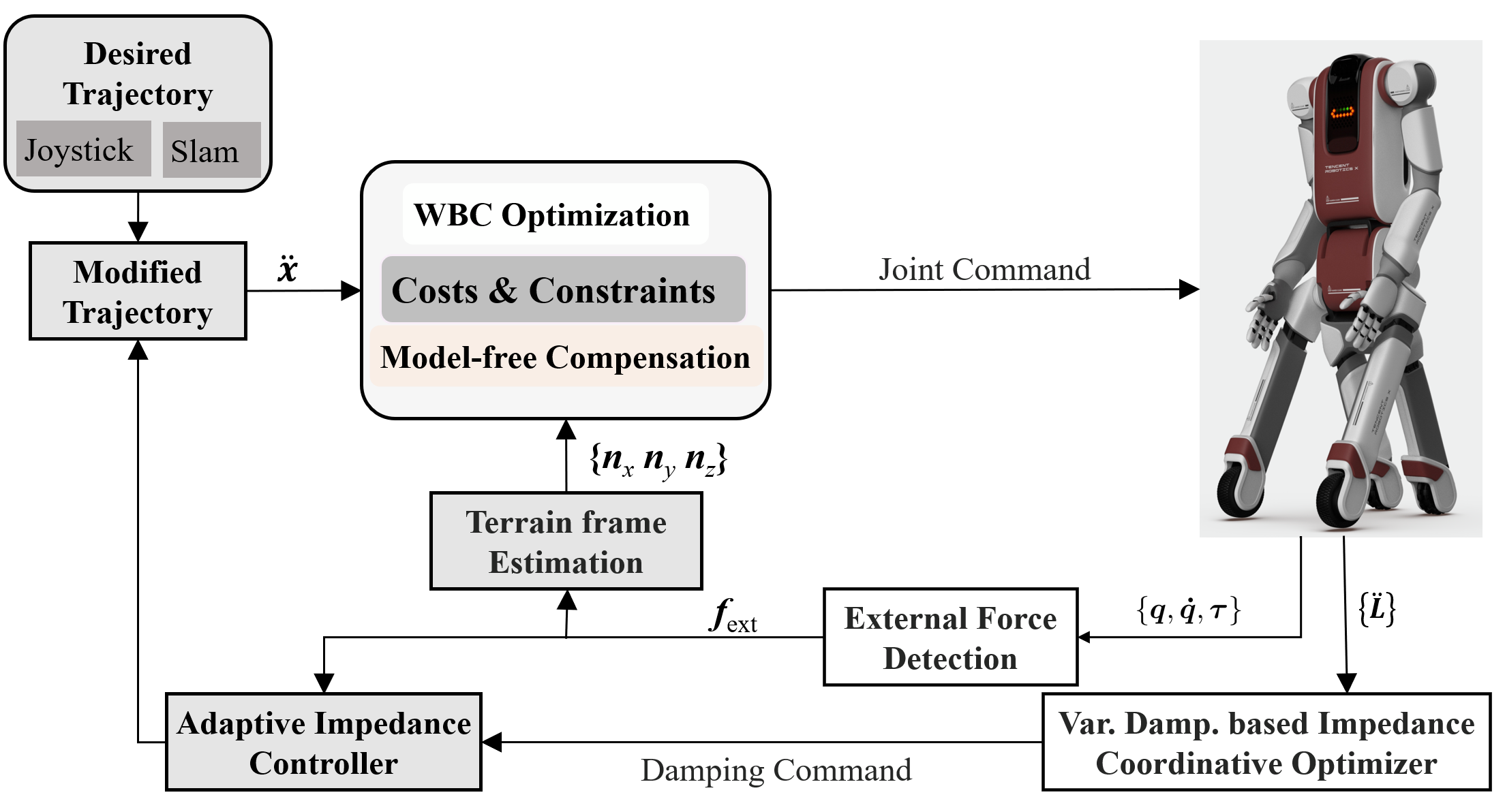}
\caption{The overview controller framework of X-Man.
Denote $\boldsymbol{q}$
% \in \mathbb{R}^{n+6}
 and $\boldsymbol{x}$ are the joint value and point state of robot.
$\boldsymbol{\dot q},\boldsymbol{\ddot q}$ are the angular velocity and acceleration.
$\boldsymbol{\tau}$ is joint torque. $\{ \boldsymbol{n}_x,\boldsymbol{n}_y,\boldsymbol{n}_z\}$ is the terrain frame. And $\boldsymbol{n}_{z}$
is unit normal vector, 
and $\boldsymbol{\boldsymbol{n}}_{x},\boldsymbol{n}_{y}$ 
are two orthonormal tangent vectors described with respect to the object frame.}
\label{figWBC}
\vspace{-0.5cm}
\end{figure}

To achieve both control objectives above, we decompose our control strategy into outer and inner loops based on the platform's architecture. The outer-loop control is to optimize  and generate optimal damping parameters. The inner-loop control, which demands a much higher control frequency, uses WBC to generate the control torques based on estimated terrain information and updated trajectory from the outer loop. %which will be not affected by the communication cycle of external sensors. 
The overall X-Man control framework, illustrated in Fig.~\ref{figWBC}, consists of five main components: the desired trajectory generation part; the outer-loop impedance control part; the inner-loop WBC part; the joint torque tracking part; and the general momentum observation part.

It is worth mentioning that since X-Man has two coupled impedance controllers in the same direction, an impedance coordinative approach is adopted to achieve the optimization goal set in the previous section~\cite{rastogi2024enhancing}.
Joint force tracking is employed to monitor and regulate the driving torques for each joint, as computed by the WBC framework. 

%
%
% \begin{figure}[htbp]

\section{Impedance Coordinative Controller}

\subsection{General Impedance  Control}

% % 一般来说，机器人的地面支持力是保证机器人和地面充分接触的，但是在实际的非结构化环境中，地形的不确定会引起机器人和地面意外的冲击碰撞，即支持力的一部分用于支持机器人与地面的基础，另一部分相对机器人来说则是扰动。 
% Generally, the ground support force of a robot is to ensure that the robot is in full contact with the ground. However, the rough terrain can cause unexpected impact and collision between the robot and the ground. That is, part of the reasonable contact force is used to support the robot on the ground, and the other part is the disturbance for the robot.

% 上述的WBC控制算法中仅仅用到了各个笛卡尔任务的位置反馈，接下来我们在WBC中引入阻抗控制，可以使得机器人主动根据力跟踪误差进行加速度修正。
% We introduce impedance control in WBC, which allows the robot to actively correct the acceleration variance based on the force tracking.
% %
% And we also propose a variable damper strategy to achieve stability in upper body manipulation.
%

The closed-loop impedance control model is governed as:
\begin{equation}
    \boldsymbol{M} \Delta \boldsymbol{\ddot{x}} + \boldsymbol{D} \Delta \boldsymbol{\dot{x}} + \boldsymbol{K} \Delta \boldsymbol{x} = \boldsymbol{F}_{e}
    \label{imped}
\end{equation}
where \( \boldsymbol{M} \), \( \boldsymbol{D} \) and \( \boldsymbol{K} \) represent the inertia, damping, and stiffness matrices, respectively. The term \( \boldsymbol{F}_{e} \) denotes the external force corresponding to the displacement \( \boldsymbol{x} \), where \( \boldsymbol{x} \) represents a specific state of robot.
Denote $\boldsymbol x^{'}_{\text{ref}}(t)$ the reference value.
$\Delta \boldsymbol {x}=\boldsymbol x(t)-\boldsymbol x^{'}_{\text{ref}}(t)$ denotes the position error.
%
% 
%
%
% Based on the impedance model in Eq.~(\ref{imped}), a modified reference acceleration \( \boldsymbol{\ddot{x}}_{\text{ref}} \) is derived as:
Then, the modified reference acceleration \( \boldsymbol{\ddot{x}}_{\text{ref}} \) is derived as:
%
% \begin{equation}
%     \boldsymbol{\ddot{x}}_{\text{ref}} = \boldsymbol{\ddot{x}}^{'}_{\text{ref}}
%     + \boldsymbol{M}^{-1} \left( \boldsymbol{F}_{e} - \boldsymbol{D} \Delta \boldsymbol{\dot{x}} - \boldsymbol{K} \Delta \boldsymbol{x} \right).
% \label{eq:imped}
% \end{equation}
\begin{equation}
    \boldsymbol{\ddot{x}} = \boldsymbol{\ddot{x}}^{'}_{\text{ref}}
    + \boldsymbol{M}^{-1} \left( \boldsymbol{F}_{e} - \boldsymbol{D} \Delta \boldsymbol{\dot{x}} - \boldsymbol{K} \Delta \boldsymbol{x} \right).
\label{eq:imped}
\end{equation}
By integrating \( \boldsymbol{\ddot{x}} \), the corresponding  updated reference velocity  \( \boldsymbol{\dot{x}}_{\text{ref}} \) and updated reference position \( \boldsymbol{x}_{\text{ref}} \) can be obtained.
%
% \( \boldsymbol{\ddot{x}}^{'}_{\text{ref}} \)
%  is pre-planned.
The updated reference values are utilized in the WBC controller.

% \subsection{Impedance Control for Ground Contact}
% \subsection{Impedance Control for Locomotion}
%
%
The locomotion-related impedance control point can be set at either the base frame (\( \Sigma_B \)) or the Center of mass frame (\( \Sigma_O \)). Accordingly, the external force must first be transformed to the chosen reference point. 
Denote
$\boldsymbol{f}_{\text{ext}}
=
[\boldsymbol{f}^T_{M} \  \boldsymbol{f}^T_{C}]^T
\in
\mathbb R^{(N_D N_C)}$  the external force.
$\boldsymbol{f}_{M}$ is the force for object manipulation.
$\boldsymbol{f}_{C}$ is the ground contact force.
$N_C$ is the number of contact point.
$N_D$ is the dimension of contact force.
For example, consider the case where the control point is set at \( \Sigma_B \), the base wrench ${\boldsymbol {\hat F}}_{e,C}$ can be derived as:
\begin{equation}
    {\boldsymbol {\hat F}}_{e}=
    \sum {\boldsymbol J^T_{B,M,i}}
    \boldsymbol S_{M,i} {\boldsymbol {\hat f}}_{\text{ext}}
    +
    \sum {\boldsymbol J^T_{B,C,i}}
    \boldsymbol S_{C,i}{\boldsymbol {\hat f}}_{\text{ext}}
    % \right
\label{eq:f_ext}
\end{equation}
where $\boldsymbol S_{M,i}, \boldsymbol S_{C,i}$ denote the section matrix for the $i$-th limb, and
$\boldsymbol S_{M,i} {\boldsymbol {\hat f}}_{\text{ext}}={\boldsymbol {\hat f}}_{M}$, 
$\boldsymbol S_{C,i} {\boldsymbol {\hat f}}_{\text{ext}}={\boldsymbol {\hat f}}_{C}$,
${\boldsymbol J_{B,M,i}}$ denotes the contact Jacobian matrix for the $i$-th limb for the manipulation task, ${\boldsymbol J_{B,C,i}}$ denotes the contact Jacobian matrix for the $i$-th limb for the locomotion task.
$\Sigma_B$ or $\Sigma_O$ will actively adjust its pose according to external forces.
${\boldsymbol {\hat f}}_{\text{ext}}$ can be obtained by the general momentum observer~\cite{focchi2020heuristic,bledt2018contact}.

\subsection{Impedance Coordinative Control}
\begin{figure}[htbp]%htbp
\captionsetup{font={footnotesize}}
\centering %
\includegraphics[width=0.5\textwidth]{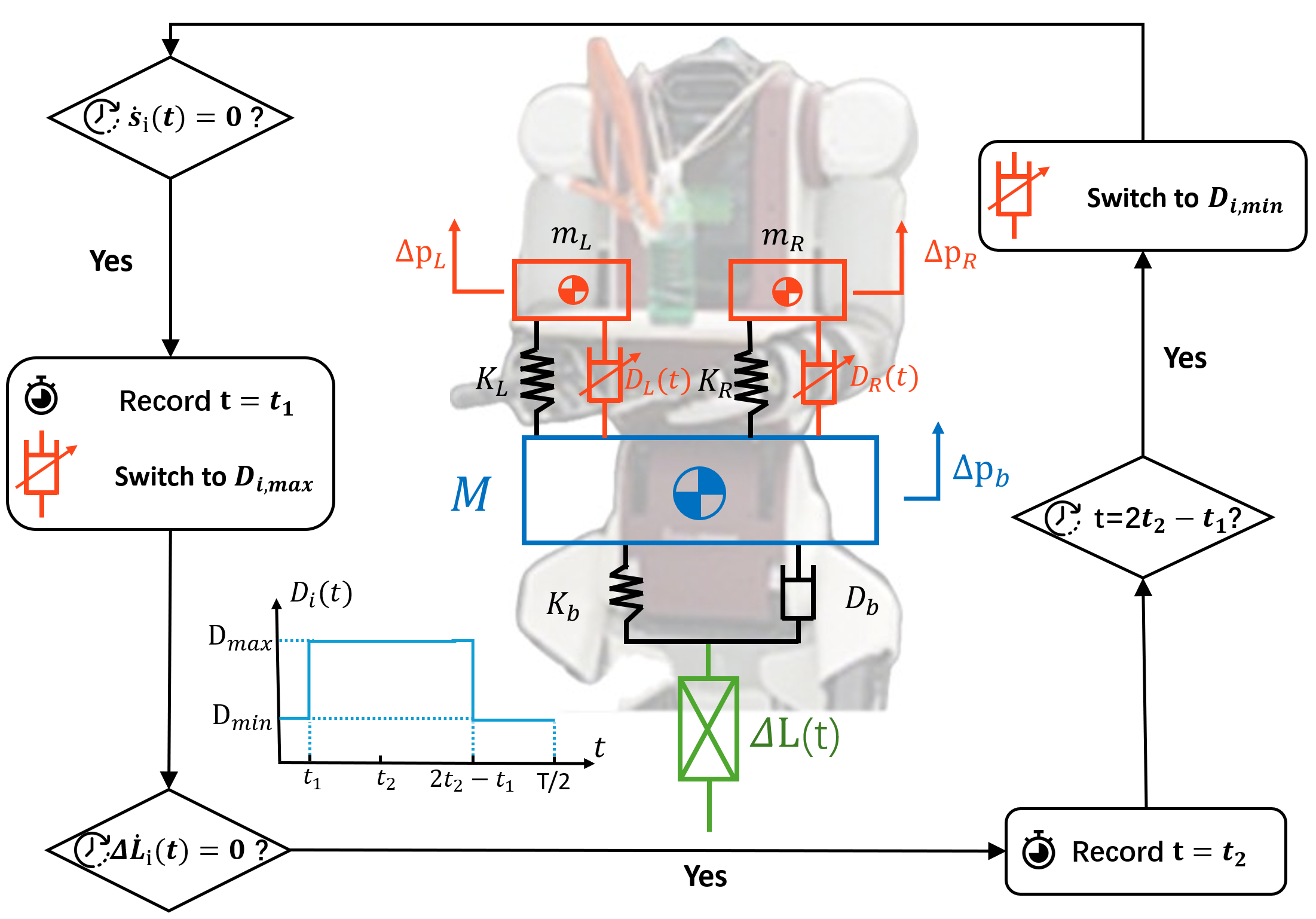}
\caption{The dual-arm variable damping-based impedance coordinative controller, where \(t_1\) represents the instant of zero relative motion between the load and the robot body, and \(t_2\) is the instant when the variation in robot leg length reaches a steady state.
The coefficients \( \boldsymbol{K}_b \), \( \boldsymbol{K}_L \), and \( \boldsymbol{K}_R \) are the stiffness matrices, and \( \boldsymbol{D}_b \), \( \boldsymbol{D}_L \), and \( \boldsymbol{D}_R \) are the damping matrices. These matrices are diagonal, representing stiffness and damping properties in the \(x\), \(y\), and \(z\) directions for the body and loads. \( M \) represents the body mass, while \( m_L \) and \( m_R \) represent the masses of the left and right loads. And  \( m_L \), \( m_R \),  and \( M \) are the components in $\boldsymbol{M}$ of~(\ref{imped}).
}
% \vspace{-0.4cm}
\label{figdampingub}
\end{figure}
To maintain upper body stability during locomotion tasks in a large inertia robot, we treat the upper arms as a coupled dynamic system. Inspired by the suspended backpack principle~\cite{YANG2020103738}, the dual arms are modeled as spring-variable damping mechanisms designed for lower limb energy saving and upper limbs passive stability. Passive stability allows the robot to keep an object stable, such as a tray, steady by naturally responding to dynamic external forces rather than actively adjusting its position.

\subsubsection{Equality constraint of ICC}

According to the mechanical structure of the X-man, we simplify the robot body\footnote{Here, the term "body" refers to the robot body, including the robot arms but excluding its lower limbs, as \(\Delta L(t)\) represents the lower limbs.}
as a mass block with a variable damper-spring mechanism connected to the left and right loads, and with an additional spring-damper mechanism below connecting with the robot legs, as shown Fig \ref{figdampingub}. We can then derive the dynamics equations of the body center of mass (CoM) and the loads.
\begin{equation}
\footnotesize
\begin{aligned}
&M \Delta\ddot{\boldsymbol{p}}_b + (\boldsymbol{D}_L + \boldsymbol{D}_L + \boldsymbol{D}_R) \Delta\dot{\boldsymbol{p}}_b + (\boldsymbol{K}_b + \boldsymbol{K}_L + \boldsymbol{K}_R) \Delta\boldsymbol{p}_b  = \\
&\boldsymbol{K}_b \Delta\boldsymbol{L} + \boldsymbol{D}_b \Delta \dot{\boldsymbol{L}} + \begin{bmatrix} \boldsymbol{D}_L & \boldsymbol{D}_R \end{bmatrix} \begin{bmatrix} \Delta\dot{\boldsymbol{p}}_L \\ \Delta\dot{\boldsymbol{p}}_R \end{bmatrix} + \begin{bmatrix} \boldsymbol{K}_L & \boldsymbol{K}_R \end{bmatrix} \begin{bmatrix} \Delta\boldsymbol{p}_L \\ \Delta\boldsymbol{p}_R \end{bmatrix}
\end{aligned}
\label{manipulation_com_dy_3d}
\end{equation}
\begin{equation}
\footnotesize
\begin{aligned}
&
\begin{bmatrix} m_L  & 0 \\ 0 & m_R \end{bmatrix}
\begin{bmatrix} \Delta\ddot{\boldsymbol{p}}_L \\ \Delta\ddot{\boldsymbol{p}}_R \end{bmatrix} 
+ \begin{bmatrix} \boldsymbol{D}_L & 0 \\ 0 & \boldsymbol{D}_R \end{bmatrix} 
\begin{bmatrix} \Delta\dot{\boldsymbol{p}}_L \\ \Delta\dot{\boldsymbol{p}}_R \end{bmatrix}
+ \begin{bmatrix} \boldsymbol{K}_L & 0 \\ 0 & \boldsymbol{K}_R \end{bmatrix} 
\begin{bmatrix} \Delta\boldsymbol{p}_L \\ \Delta\boldsymbol{p}_R \end{bmatrix} \\
& = 
\begin{bmatrix} \boldsymbol{D}_L & 0 \\ 0 & \boldsymbol{D}_R \end{bmatrix}
\begin{bmatrix} \Delta\dot{\boldsymbol{p}}_b \\ \Delta\dot{\boldsymbol{p}}_b \end{bmatrix} 
+ \begin{bmatrix} \boldsymbol{K}_L & 0 \\ 0 & \boldsymbol{K}_R \end{bmatrix}
\begin{bmatrix} \Delta\boldsymbol{p}_b \\ \Delta\boldsymbol{p}_b \end{bmatrix} 
\end{aligned}
\label{manipulation_load_dy_3d}
\end{equation}
Here, \( \Delta\boldsymbol{p}_b = [\Delta p_{b_x}, \Delta p_{b_y}, \Delta p_{b_z}]^T \) represents the variation in the position of the body CoM in the \(x\), \(y\), and \(z\) directions, while \( \Delta\boldsymbol{p}_L = [\Delta p_{L_x}, \Delta p_{L_y}, \Delta p_{L_z}]^T \) and \( \Delta\boldsymbol{p}_R = [\Delta p_{R_x}, \Delta p_{R_y}, \Delta p_{R_z}]^T \) denote the variations in the positions of the left and right loads, respectively. Other notations are defined in Fig. \ref{figdampingub} and elaborated in the following two remarks.

\begin{remark}
The values of \(m_L\) and \(m_R\) are determined based on the gravity measured by the tactile skin in a static state. These loads can either represent two different objects held by the left and right upper limbs or a single object with its weight distributed between both limbs.
\end{remark}

\begin{remark}
The function \(\Delta\ddot{\boldsymbol{L}}(t) \in \mathbb{R}^3\) is determined by leg-hip motion, which can be pre-defined through motion planning and measured in real time using IMU sensors. Note that \(\Delta\boldsymbol{L}(t)\) represents the variation in leg length during motion, not the absolute leg length.
\(\Delta\boldsymbol{L}(t)\) will directly determine the movement of the robot base and indirectly cause the change of the robot's Center of mass.
Moreover, \(\Delta \boldsymbol{L}(t)\) arises as a response to dynamic external contact forces acting on the lower limbs. These contact forces introduce oscillations that propagate through the robot's structure, necessitating passive stability to mitigate any impact on upper body.
\end{remark}

By defining the relative displacements as \( \boldsymbol{s_b} = \Delta\boldsymbol{p}_b - \Delta\boldsymbol{L} \), \( \boldsymbol{s_L} = \Delta\boldsymbol{p}_L - \Delta\boldsymbol{p}_b \), \( \boldsymbol{s_R} = \Delta\boldsymbol{p}_R - \Delta\boldsymbol{p}_b \), the system equations can be reformulated. Let the state vector be defined as:
\[
\boldsymbol{x_s} = \begin{bmatrix} \boldsymbol{s_b} & \dot{\boldsymbol{s}}_b & \boldsymbol{s_L} & \dot{\boldsymbol{s}}_L & \boldsymbol{s_R} & \dot{\boldsymbol{s}}_R \end{bmatrix}^T,
\]
then the system dynamics can be expressed as:

\begin{equation}
\dot{\boldsymbol x}_s = \boldsymbol{A}(t) \boldsymbol {x_s} + \boldsymbol{b}(t)
\label{eq:CoM Dynamics}
\end{equation}
where the matrix \(\boldsymbol{A}(t)\) and input vector \(\boldsymbol{b}(t)\) are defined as:

\[
\footnotesize
\boldsymbol {A} = 
\left[\begin{smallmatrix}
0 & \mathbf{I_{3\times3}} & 0 & 0 & 0 & 0 \\
-M^{\!-1} \boldsymbol{K}_b & -M^{\!-1} \boldsymbol{D}_b & M^{\!-1} \boldsymbol{K}_L & M^{\!-1} \boldsymbol{D}_L & M^{\!-1} \boldsymbol{K}_R & M^{\!-1} \boldsymbol{D}_R \\
0 & 0 & 0 & \mathbf{I_{3\times3}} & 0 & 0 \\
M^{\!-1} \boldsymbol{K}_b & M^{\!-1} \boldsymbol{D}_b & -\boldsymbol{K}_L \boldsymbol{C_L} & -\boldsymbol{D}_L \boldsymbol{C_L} & -M^{\!-1} \boldsymbol{K}_R & -M^{\!-1} \boldsymbol{D}_R \\
0 & 0 & 0 & 0 & 0 & \mathbf{I_{3\times3}} \\
M^{\!-1} \boldsymbol{K}_b & M^{\!-1} \boldsymbol{D}_b & -M^{\!-1} \boldsymbol{K}_L & -M^{\!-1} \boldsymbol{D}_L & -\boldsymbol{K}_R \boldsymbol{C}_R & -\boldsymbol{D}_R \boldsymbol{C}_R
\end{smallmatrix} \right]
\]

\[
% \footnotesize
\boldsymbol b(t) = 
\begin{bmatrix}
0 & -\Delta\ddot{\boldsymbol{L}} & 0 & 0 & 0 & 0
\end{bmatrix}^T
\]
The matrices \(\boldsymbol{D}_L\) and \(\boldsymbol{D}_R\) serve as control inputs. The coefficients \(\boldsymbol{C}_L = {m_L}^{\!-1} + {M}^{\!-1}\) and \(\boldsymbol{C}_R = {m_R}^{\!-1} + {M}^{\!-1}\) account for the body and load masses. The function \(\Delta \boldsymbol{L}(t)\) represents the effective length variation of the lower limbs.

\subsubsection{Cost Function of ICC}

% The energy expenditure for manipulating the load is calculated by integrating the mechanical power \(P_1(t)\) over a cycle \(T\).
% Thus, the net mechanical work is given by:
For the locomotion task, the mechanical power can be derived as:
% net mechanical work of lower limbs is given by:
%
% \begin{equation}
% \footnotesize
% \begin{aligned}
% E_1 &= \int_0^T \rho_1(t) \, dt \\
% &= \int_0^T \left[ -\boldsymbol{K}_b \boldsymbol {s_b} - \boldsymbol{D}_b \dot{\boldsymbol {s}}_{\boldsymbol b} + \begin{pmatrix}
% 0 \\
% 0 \\
% (M + m_L + m_R) \boldsymbol{g}
% \end{pmatrix}
% \right]^T \Delta\dot{\boldsymbol{L}} \, dt
% \end{aligned}
% \end{equation}
%
%
\begin{equation}
% \footnotesize
\begin{aligned}
\rho_1(t)= \left[ -\boldsymbol{K}_b \boldsymbol {s_b} - \boldsymbol{D}_b \dot{\boldsymbol {s}}_{\boldsymbol b} + \begin{pmatrix}
0 \\
0 \\
(M + m_L + m_R) \boldsymbol{g}
\end{pmatrix}
\right]^T \Delta\dot{\boldsymbol{L}} \
\end{aligned}
\end{equation}

If \(\Delta\dot{L}(t) = 0\), as in motion on flat terrain where the robot's body CoM shows no oscillation, then \(E_1 = 0\). However, if \(\Delta\dot{L}(t) \neq 0\), the oscillation of the body CoM will cause the lower limbs to expend additional energy, which can be significant for robots with large inertia, such as our 80 kg X-man robot.
Then we can derive the mechanical power $\rho_2(t)$ for the passive stability as
% We introduce a weighted term for the passive stability in the cost function:
%
%
% \begin{equation}
% % \footnotesize
% \begin{aligned}
% E_2 &= \int_0^T \rho_2(t) \, dt \\
% &= \int_0^T \frac{1}{2}\left( 
%     \dot{\boldsymbol{s}}_L^T \mathbf{W}_L \dot{\boldsymbol{s}}_L + 
%     \dot{\boldsymbol{s}}_R^T \mathbf{W}_R \dot{\boldsymbol{s}}_R 
%     \right) \, dt,
% \end{aligned}
% \end{equation}
%
%
\begin{equation}
% \footnotesize
\begin{aligned}
\rho_2(t)=   \frac{1}{2}\left( 
    \dot{\boldsymbol{s}}_L^T \mathbf{W}_L \dot{\boldsymbol{s}}_L + 
    \dot{\boldsymbol{s}}_R^T \mathbf{W}_R \dot{\boldsymbol{s}}_R 
    \right) \,
\end{aligned}
\end{equation}
The symmetric positive-definite weight matrices \( \mathbf{W}_L \) and \( \mathbf{W}_R \) allow for anisotropic penalization across different spatial directions, reflecting the specific stability requirements of each limb. By minimizing this energy term, the system discourages rapid changes in limb motion, ensuring controlled and coordinated behavior throughout the task duration \( T \).

\subsubsection{Optimization for ICC}

The optimal control problem for dual-arm in three dimension is defined as:
% \begin{equation}
% \footnotesize
% \begin{aligned}
% \min E &= \int_0^T \left[ \left( -\boldsymbol{K}_b \boldsymbol {s_b} - \boldsymbol{D}_b \dot{\boldsymbol {s}}_{\boldsymbol b}\right)^T \Delta\dot{\boldsymbol{L}} +  \frac{1}{2}\left( 
%     \dot{\boldsymbol{s}}_L^T \mathbf{W}_L \dot{\boldsymbol{s}}_L + 
%     \dot{\boldsymbol{s}}_R^T \mathbf{W}_R \dot{\boldsymbol{s}}_R 
%     \right)
%     \right] dt \\
% \text{subject to} \quad &\dot{\boldsymbol{x}}_s(t) = \boldsymbol A(t)\boldsymbol{x_s}(t) + \boldsymbol b(t), \\
% &\boldsymbol{x_s}(0) = \boldsymbol{x_s}(T), \\
% & \Delta\boldsymbol{D}_L(t) \in [\Delta D_{\text{L,min}}, \Delta D_{\text{L,max}}],\\
% & \Delta\boldsymbol{D}_R(t) \in [\Delta D_{\text{R,min}}, \Delta D_{\text{R,max}}].
% \end{aligned}
% \end{equation}
\begin{equation}
\footnotesize
\begin{aligned}
\min E &= \int_0^T  (\rho_1(t)+\rho_2(t)) dt \\
\text{subject to} \quad &\dot{\boldsymbol{x}}_s(t) = \boldsymbol A(t)\boldsymbol{x_s}(t) + \boldsymbol b(t), \\
&\boldsymbol{x_s}(0) = \boldsymbol{x_s}(T), \\
& \Delta\boldsymbol{D}_L(t) \in [\Delta D_{\text{L,min}}, \Delta D_{\text{L,max}}],\\
& \Delta\boldsymbol{D}_R(t) \in [\Delta D_{\text{R,min}}, \Delta D_{\text{R,max}}].
\end{aligned}
\end{equation}
where the subscripts ‘$\text{min}$’ and ‘$\text{max}$’ represent the minimum and maximum values, respectively.

The Hamiltonian for this problem is expressed as:
\begin{equation}
\footnotesize
\begin{aligned}
H &= \left[ -\boldsymbol{K}_b \boldsymbol {s_b}(t) - \boldsymbol{D}_b \dot{\boldsymbol {s}}_{\boldsymbol b}(t) \right]^T \Delta\dot{\boldsymbol{L}}(t) + 
    \frac{1}{2}\left( 
    \dot{\boldsymbol{s}}_L^T \mathbf{W}_L \dot{\boldsymbol{s}}_L + 
    \dot{\boldsymbol{s}}_R^T \mathbf{W}_R \dot{\boldsymbol{s}}_R 
    \right) \\
&\quad + \boldsymbol{\lambda}^T(t)\left[\boldsymbol A(t){\boldsymbol x}_s(t) + \boldsymbol b(t)\right] 
% \right)
\end{aligned}
\end{equation}
where \(\boldsymbol{\lambda}(t)\) is Lagrange multiplier, satisfying the co-state equation:
\begin{equation}
% \footnotesize
\begin{aligned}
    \dot{\boldsymbol{\lambda}}(t) &= -\frac{\partial H}{\partial \boldsymbol{x}_s}= -A^T(t) \boldsymbol{\lambda}(t) + \begin{bmatrix} \boldsymbol{K}_b^T \Delta\dot{\boldsymbol{L}}(t) \\ \boldsymbol{D}_b^T \Delta\dot{\boldsymbol{L}}(t) \\ 0 \\ -\mathbf{W}_L \dot{\boldsymbol{s}}_L \\ 0 \\ -\mathbf{W}_R \dot{\boldsymbol{s}}_R \end{bmatrix}
\end{aligned}
\end{equation}

Specifically, we have:
\begin{equation}
% \footnotesize
\begin{aligned}
&\dot{\boldsymbol{\lambda}}_3(t) = -\boldsymbol{K}_L \left[ M^{\!-1}\boldsymbol{\lambda}_2  - \left( M^{\!-1} + m_L^{\!-1} \right) \boldsymbol{\lambda}_4  - M^{\!-1}\boldsymbol{\lambda}_6\right] \\
&\dot{\boldsymbol{\lambda}}_5(t) = -\boldsymbol{K}_R \left[ M^{\!-1}\boldsymbol{\lambda}_2  - M^{\!-1}\boldsymbol{\lambda}_4 - \left( M^{\!-1} + m_R^{\!-1} \right) \boldsymbol{\lambda}_6 \right] 
\end{aligned}
\end{equation}

Substituting this expression into the Hamiltonian yields:
\begin{equation}
\begin{aligned}
H &= H_0[{\boldsymbol x}_s(t), \boldsymbol{\lambda}(t), t] - \dot{\boldsymbol{s}}_{\boldsymbol{L}}^T(t) \boldsymbol{D}_L(t) {\boldsymbol{K}_L}^{\!-1} \dot{\boldsymbol{\lambda}}_3(t) \\
&\quad - \dot{\boldsymbol{s}}_{\boldsymbol{R}}^T(t) \boldsymbol{D}_R(t) {\boldsymbol{K}_R}^{\!-1} \dot{\boldsymbol{\lambda}}_5(t)
\end{aligned}
\end{equation}

Since \(H\) is a scalar function that encapsulates contributions from all three dimensions (i.e., \(H = H_x + H_y + H_z\)), and our control input—the variable dampers \(\boldsymbol{D}_L\) and \(\boldsymbol{D}_R\)—is also three-dimensional, we need to consider each dimension \(i\) (where \(i = x, y, z\)) separately. According to Pontryagin's Minimum Principle, we apply bang-bang control to the damping terms \( \boldsymbol{D}_L \) and \( \boldsymbol{D}_R \) for each dimension \(i\):

\begin{equation}
\begin{aligned}
&\boldsymbol{D}_{L,i}(t) = 
\begin{cases} 
D_{L,i,\text{min}}, & \text{if } \dot{\boldsymbol{s}}_{L,i}^T(t) {\boldsymbol{K}_{L,i}}^{\!-1} \dot{\boldsymbol{\lambda}}_{3,i}(t) < 0, \\
D_{L,i,\text{max}}, & \text{if } \dot{\boldsymbol{s}}_{L,i}^T(t) {\boldsymbol{K}_{L,i}}^{\!-1} \dot{\boldsymbol{\lambda}}_{3,i}(t) > 0,
\end{cases}  \\
&\boldsymbol{D}_{R,i}(t) = 
\begin{cases} 
D_{R,i,\text{min}}, & \text{if } \dot{\boldsymbol{s}}_{R,i}^T(t) {\boldsymbol{K}_{R,i}}^{\!-1} \dot{\boldsymbol{\lambda}}_{5,i}(t) < 0, \\
D_{R,i,\text{max}}, & \text{if } \dot{\boldsymbol{s}}_{R,i}^T(t) {\boldsymbol{K}_{R,i}}^{\!-1} \dot{\boldsymbol{\lambda}}_{5,i}(t) > 0,
\end{cases}  
\end{aligned}
\end{equation}

% \begin{remark}
% \(\boldsymbol{D}_L\) is a \(3 \times 3\) diagonal matrix, so \( \boldsymbol{D}_{L,\text{min}} \) can be written as \(\text{diag}(D_{L_{xx},\text{min}}, D_{L_{yy},\text{min}}, D_{L_{zz},\text{min}})\). The same applies to \(\boldsymbol{D}_R\).
% \end{remark}

% \begin{equation}
% \begin{aligned}
% &\boldsymbol{D}_L^*(t) = 
% \begin{cases} 
% D_{L,\text{min}}, & \text{if } \dot{\boldsymbol {s}}_{\boldsymbol L}(t) \cdot \dot{\boldsymbol{\lambda}}_3(t) < 0, \\
% D_{L,\text{max}}, & \text{if } \dot{\boldsymbol {s}}_{\boldsymbol L}(t) \cdot \dot{\boldsymbol{\lambda}}_3(t) > 0,
% \end{cases}  \\
% &\boldsymbol{D}_R^*(t) = 
% \begin{cases} 
% D_{R,\text{min}}, & \text{if } \dot{\boldsymbol {s}}_{\boldsymbol R}(t) \cdot \dot{\boldsymbol{\lambda}}_5(t) < 0, \\
% D_{R,\text{max}}, & \text{if } \dot{\boldsymbol {s}}_{\boldsymbol R}(t) \cdot \dot{\boldsymbol{\lambda}}_5(t) > 0.
% \end{cases}   
% \end{aligned}
% \end{equation}

The elements of the diagonal matrices \(\boldsymbol{K}_L\) and \(\boldsymbol{K}_R\) are non-negative, which does not affect the positive or negative sign. Although \(\dot{s}_{L,i}\) can be observed directly from its physical interpretation, obtaining \(\dot{\lambda}_{3,i}(t)\) is not as straightforward. To implement the bang-bang control strategy, it is necessary to identify the time at which \(\dot{\lambda}_{3,i}(t) = 0\). As demonstrated in \cite{YANG2020103738}, the variable damper switches precisely at the zero-crossing of \(\dot{s}_{X,i}(t) \quad \text{where } X \in \{L, R\}\), symmetrically aligned with the zero-crossing of \( \dot{s}_{X,i}(t)\) relative to \(\Delta\dot{L}_i(t)\). Let \(t_1\) denote the time at which \(\dot{s}_{L,i} = 0\), and \(t_2\) the time at which \(\Delta\dot{L} = 0\); then the time when \(\dot{\lambda}_{3,i} = 0\) is given by \(2t_2 - t_1\). A similar relationship applies to \(\dot{\lambda}_{5,i}(t)\) and \(\dot{\boldsymbol s}_{R,i}(t)\).

In Equ. \eqref{eq:CoM Dynamics}, the lower limb trajectory is assumed to be already obtained. However, during dual-arm manipulation, coupling effects emerge due to the variable damping in the arms, which directly influences the dynamics of the body’s CoM. This transmits the effects to the lower limbs, further impacting their dynamics. The coupling force between the upper arms and the CoM is modeled as:

\begin{equation*}
    \boldsymbol F_{\text{cpl}} =  \begin{bmatrix} \boldsymbol{D}_L & \boldsymbol{D}_R \end{bmatrix} \begin{bmatrix} \Delta\dot{\boldsymbol{p}}_L \\ \Delta\dot{\boldsymbol{p}}_R \end{bmatrix} + \begin{bmatrix} \boldsymbol{K}_L & \boldsymbol{K}_R \end{bmatrix} \begin{bmatrix} \Delta\boldsymbol{p}_L \\ \Delta\boldsymbol{p}_R \end{bmatrix},
\end{equation*}

The disturbance force, \( \boldsymbol F_{\text{cpl}} \), along with the load's gravity, can be detected through tactile sensors and incorporated into the external force term, \( \boldsymbol{\hat f}_{\text{ext}} \), in Eq. \eqref{eq:f_ext}.    

% ~~~~~~~~~~~~~~~~~~~~~~~~~~~~~~~~~~~~~~~~~~~~~~~~~~~~~~~~~~~~~~~~~~~`

\section{Whole Body Control of X-Man}
 We can obtain the extended task $\boldsymbol{x}=[\boldsymbol{x}_R^T \  \boldsymbol{x}^T_C]^T$, and extended Jacobian $\boldsymbol{J}=[\boldsymbol{J}_R^T \  
\boldsymbol{J}^T_C]^T$.
 Denote $\boldsymbol{x}_R=\boldsymbol{f}(\boldsymbol{q})$ the task-space task. 
$\boldsymbol{J}_R=\frac{\partial \boldsymbol{f}}{\partial  \boldsymbol{q}}$, denotes the jacobian matrix. 
 $\boldsymbol{J}_C$ is the contact Jacobian matrix.  $\boldsymbol{\ddot x}_C$ denotes the Cartesian 
 acceleration in the constraint direction of the contact points. For a fixed contact point,  
$\boldsymbol{\ddot x}_C=\boldsymbol 0$. 
So the second-order differential of  $\boldsymbol{x}_{\kappa}, {\kappa}=R,C$  can be shown as
\begin{equation} \label{con:kinematics-whole-robot}
\boldsymbol{\ddot x}_{\kappa}=\boldsymbol{ J}_{\kappa}\boldsymbol{\ddot q}+\boldsymbol{\dot J}_{\kappa}\boldsymbol{\dot q} 
\end{equation}
The floating  base robot's dynamics can be described as 
\begin{equation} \label{con:dynamics-whole-robot}
\boldsymbol{B}(\boldsymbol q)\boldsymbol{\ddot q}+ \mathbf{C}(\boldsymbol{q,\dot q)}=\boldsymbol S^T (\boldsymbol{\tau}+\boldsymbol{\tau}_f)
+
\boldsymbol{J}^T_{\text{e}}\boldsymbol{f}_{\text{ext}}  
\end{equation}
where $\boldsymbol B \in \mathbb{R}^{(n+6) \times (n+6)}$
denotes the inertia matrix.
$\mathbf{C}\in \mathbb{R}^{(n+6)}$ is the vector of  bias force. 
$\boldsymbol{S}\in \mathbb{R}^{n \times (n+6)}$ is the selection matrix. 
%
% And if the DOF of free-floating base is $N_F$, 
%
% ${\boldsymbol{S}=[0_{n\times 6} \ 
% 1_{n\times n}
% ]^T}$.
%
$\boldsymbol{\tau}\in \mathbb R^{n}$ is the vector of joint driving torques.
$\boldsymbol{\tau}_f$ is the vector of joint nonlinear torques come from friction, communication delay, and mechanical dead zone.
%
% The robot body will have a nonlinear disturbance $\boldsymbol{\tau}_f$ due to factors such as joint friction, communication delay, and mechanical dead zone.
% $\boldsymbol{f}_{\text{ext}}
% =
% [\boldsymbol{f}^T_{M} \  \boldsymbol{f}^T_{C}]^T
% \in
% \mathbb R^{(N_D N_C)}$ is the external torques.
% %
% $\boldsymbol{f}_{M}$ is the force for object manipulation.
% %
% $\boldsymbol{f}_{C}$ is the ground contact force.
%
%
$\boldsymbol{J}_{\text{e}}
\in
\mathbb R^{{(N_D N_C)} \times (n+6)}$ is the Jacobian matrix.
%
% $N_G=N_F+N_J$ is the DOFs of robot.
%
% $N_C$ is the number of contact point.
% %
% %
% $N_D$ is the dimension of contact force.

\subsection{Formulation of Whole Body Control}
\subsubsection{Equality Constraint}
 % The position-velocity feedback controller can be utilized to update the desired acceleration $\boldsymbol{\ddot x}_{\text{des}}$.
  The position-velocity feedback controller can be utilized to can be derived as.
\begin{equation} \label{con:dynamics}
\boldsymbol{\ddot x}_{\text{des}}=\boldsymbol{\ddot x}_{\text{ref}}
+\boldsymbol{K}_P(\boldsymbol{x}_{\text{ref }}-\boldsymbol{x})
+\boldsymbol{K}_D(\boldsymbol{\dot x}_{\text{ref }}-\boldsymbol{\dot x}) 
\end{equation}
where the \(\boldsymbol{x}_{\text{ref }}\) is obtained from \eqref{eq:imped}. '\text{des}' denotes the desired value. 
$\boldsymbol K_P$ and $\boldsymbol K_D$ are the 
proportional and differential parameters.
%
%
%
%
% Contact point confusion!!!!~~~~~~~~~~~~~~~~~~~~~
%
%
Let $\boldsymbol{X}=[\boldsymbol{\ddot q^T} \; \boldsymbol{\tau}^T \; \boldsymbol{f}^T_C]^T$ 
be the quantity to be solved. From  (\ref{con:kinematics-whole-robot}) and (\ref{con:dynamics-whole-robot}), we can derive the equality constraints 
in the WBC controller.
\begin{equation} \label{con:wbcdynamics}
\boldsymbol{ZX}
=\boldsymbol{N}
\end{equation}
and
 \begin{equation} \boldsymbol{Z}=
\begin{bmatrix}
\boldsymbol{B}&
-\boldsymbol{S}^T&
-\boldsymbol{J}_e^T\\
\boldsymbol{J}&
\boldsymbol{0}&
\boldsymbol{0}
\end{bmatrix},
\ \ 
\boldsymbol{N}=
\begin{bmatrix}
-\boldsymbol{C}+\boldsymbol{S}^T\boldsymbol{\tau}_f  \\
\boldsymbol{\ddot x_{\text{des}}}-\boldsymbol{\dot J}\boldsymbol{\dot q}
\end{bmatrix}
.
% \right
\end{equation}
%
%
%
% 
%
% Here, $\boldsymbol{A}$ and $\boldsymbol{B}$ can be obtained based on the robot state and planning value.

\subsubsection{Contact Constraints}
Denote ${\mathcal{F}_i}$ the friction cone of $i$-th contact point~\cite{zhou2023differential}. ${\mathcal{F}_i}$ can be derived as: 
% The friction cone constraint is to ensure the no-slip condition.~\cite{zhou2023differential,jenelten2022tamols}.
 %
%
\begin{equation}\
  \boldsymbol{f}_{C,i}
   \in
   {\mathcal{F}_i}, i=1,2,\cdots,N
\end{equation}
% 
% Here, ${\mathcal{F}_i}$ denotes the friction cone.

\begin{equation}
  \mathcal{F}_i =
   \left\{
   \boldsymbol{f}_{C,i} \mid
   \mu_i \boldsymbol{n}^T_{z,i} \boldsymbol{f}_{C,i}
   \geq
   \lVert (\boldsymbol{I} - \boldsymbol{n}_{z,i} \boldsymbol{n}^T_{z,i}) \boldsymbol{f}_{C,i} \rVert
   \right\}
\end{equation}

\subsubsection{Optimization for WBC and Impedance Control}
% 
% The cost function  include the regularization term $\boldsymbol{X}^T\boldsymbol{X}$,
% %
% % 
%  body vibration suppression term  $\boldsymbol{\dot \tau}^T \boldsymbol{\dot \tau}$, and $\boldsymbol{\dot f}_C^T \boldsymbol{\dot f}_C$.
 %
 %
% 
%
So the whole body dynamics optimization controller integrating impedance control can be expressed as follows:
\begin{equation}
\begin{aligned}
    \mathrm{min}_{\boldsymbol X}\quad     &      
    \lVert  \boldsymbol X\lVert^2_{\boldsymbol R_X}
    +
    \lVert  {\boldsymbol {\dot \tau}}\lVert^2_{\boldsymbol R_{\tau}}
    +
    \lVert  {\boldsymbol {\dot f}}_C \lVert^2_{\boldsymbol R_{f}} \\
    \mathrm{subject \ to}\quad &\text{(\ref{con:wbcdynamics})}
    \\
     &  \boldsymbol {\tau}_{\text{min}} \leq \boldsymbol {\tau} \leq  \boldsymbol {\tau}_{\text{max}}
    \\
     &  \boldsymbol {f}_{\text{C,min}} \leq \boldsymbol {f}_C \leq  \boldsymbol {f}_{\text{C,max}}
\end{aligned}
% \right
\end{equation}
%
%
% \begin{equation}
% % \footnotesize
% \begin{aligned}
% \min E &= \int_0^T \left[ -\boldsymbol{K}_b \boldsymbol {s_b} - D_b \dot{\boldsymbol {s}}_{\boldsymbol b}  \right] \dot{\boldsymbol{L}} \, dt, \\
% \text{subject to} \quad &\dot{\boldsymbol{x}}_s(t) = A(t)\boldsymbol{x_s}(t) + b(t), \\
% &\boldsymbol{x_s}(0) = \boldsymbol{x_s}(T), \\
% & \boldsymbol{D}_L(t) \in [D_{\text{L,min}}, D_{\text{L,max}}] \\
% & \boldsymbol{D}_R(t) \in [D_{\text{R,min}}, D_{\text{R,max}}]
% \end{aligned}
% \end{equation}
%
where ${\boldsymbol R}_X$, ${\boldsymbol R}_{\tau}$ and ${\boldsymbol R}_{f}$ are the weight matrixs. Thus, the optimal value  $\boldsymbol \tau^*$ in $\boldsymbol X^*$ can be derived to control robot directly. 
%
% The subscripts ‘$\text{min}$’ and ‘$\text{max}$’ represent the minimum and maximum values, respectively.
%
Since vibration suppression terms $\lVert {\boldsymbol {\dot \tau}}\lVert $ and $\lVert {\boldsymbol {\dot f}}\lVert $ are not included in the system state equation, it can be replaced by $\lVert {\boldsymbol {\tau}-\boldsymbol {\tau}({t-\Delta t})}\lVert$ and
$\lVert {\boldsymbol {f}_C-\boldsymbol {f}_C({t-\Delta t})}\lVert $ 
in the controller, and $\Delta t$ is the control cycle.
%
 % 值得一提的是，上述人形机器人的全身动力学控制在处理任务优先级时，既可以采用加权的方式也可以采用分层的方式，具体的求解方式在其他文献中均有过充分研究，本文不再赘述。根据机器人的软硬件的研发进度，本文实验部分采用了基于权重的方式。
It is worth mentioning that WBC can use either WQP or HQP~\cite{hutter2014quadrupedal}.
% a weighted approach or a strict hierarchical approach~\cite{hutter2014quadrupedal}. 
% The specific solution methods have been fully studied  and will not be repeated in this article. 
According to the development progress of our robot, the experimental part of this article adopts a weight-based approach.

\subsection{Terrain Frame Estimation}
\begin{figure}[htbp]
\captionsetup{font={footnotesize}}
\centering %
\includegraphics[width=0.4\textwidth]{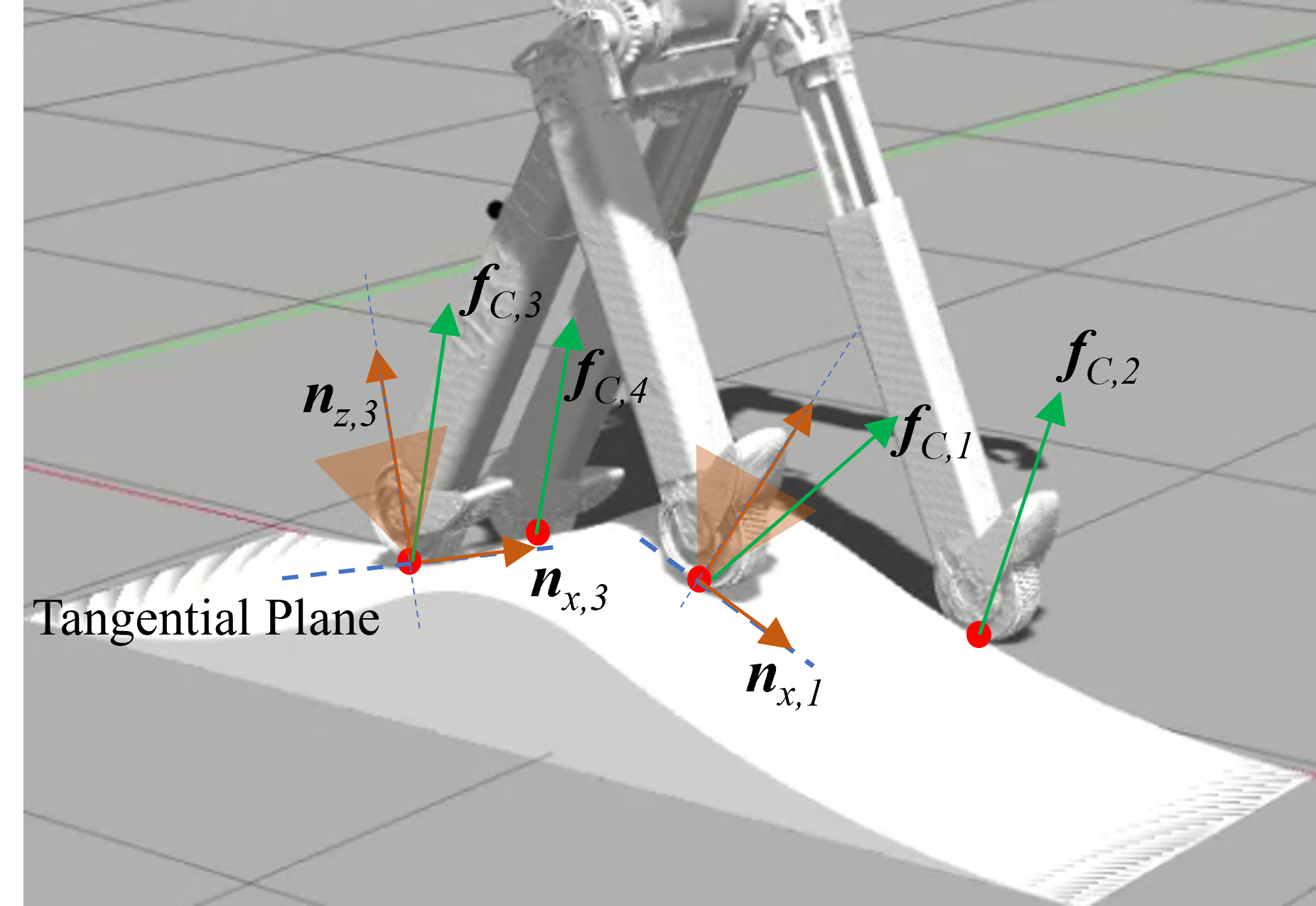}
\caption{Terrain frame estimation. $\{ \boldsymbol{n}_{x,i},\boldsymbol{n}_{y,i},\boldsymbol{n}_{z,i}\}$ is the terrain frame of $i$-th contact point. $\boldsymbol{f}_{C,i}$ is the ground contact force of $i$-th contact point.} 
\vspace{-0.4cm}
\label{terrain}
\end{figure}

This part we propose a novel  strategy for the  estimation of terrain frame $\{\boldsymbol{n}_x, \boldsymbol{n}_y$,$\boldsymbol{n}_z\}$ shown in Fig.~\ref{terrain}.
$\boldsymbol{n}_x$ can be estimated directly from the wheel's history trajectory in the forward direction, and $\boldsymbol{n}_z$ can be estimated from contact force of wheel-foot, because the force in this direction is more obviously affected by the robot's own gravity.
The constraint surface~\cite{pliego2015adaptive,namvar2005adaptive} 
% , which can be obtained by fitting the wheel history trajectory,
in the task space can be expressed as follows:
\begin{equation}
\boldsymbol \psi(\boldsymbol x,\theta)=\boldsymbol 0
% \right
\end{equation}
where $\boldsymbol \psi \in \mathbb R^m$.
% , and it can also expressed in the joint space as, $\boldsymbol \psi(\boldsymbol q,\theta)=\boldsymbol 0$.
%
%
% 
%
 Denote $\boldsymbol J_{\psi,x}(\boldsymbol x,\theta)$  the constraint Jacobian matrix in the task space, $\boldsymbol J_{\psi,x}(\boldsymbol x,\theta)
\boldsymbol {\dot x}
=\boldsymbol 0$, and it also belongs to orthogonal matrix.
\begin{equation}
\boldsymbol J_{\psi,x}(\boldsymbol x,\theta)
=\lVert\frac{\partial \boldsymbol \psi(\boldsymbol x,\theta)}
{\partial {\boldsymbol x}} \lVert ^{-1}
\frac{\partial \boldsymbol \psi(\boldsymbol x,\theta)}
{\partial {\boldsymbol x}}
% \right
\end{equation}
%
%
%
%
%
% Let $S$ the tangential plane to the constraining surface.  $S$ specifies the null of $\boldsymbol J_{\psi,x}$. And $S_T$ is orthogonal complement of $S$.
% So we can derive the orthogonal projection matrix $\boldsymbol P(\boldsymbol x,\theta)$ onto $S$. 
%
%
% 
Denote $\boldsymbol P(\boldsymbol x,\theta)$ the orthogonal projection matrix.
Let $\boldsymbol{\dot x}_{\text{forward}}$ be the forward speed, and it is also an element of $\boldsymbol{\dot x}_R$.
So we have:
\begin{equation}
\boldsymbol{P}(\boldsymbol x,\theta)= \boldsymbol I-\boldsymbol J^+_{\psi,x}(\boldsymbol x,\theta)
\boldsymbol J_{\psi,x}(\boldsymbol x,\theta)
% 
% \right
\end{equation}
\begin{equation}
 \boldsymbol {\hat n}_x=
\boldsymbol{P}  \boldsymbol {\dot x}_{\text{forward}}{\lVert\boldsymbol {\dot x}_{\text{forward}} \lVert^{-1}}
% 
% \right
\end{equation}
%
 
%
% $\boldsymbol P$ can project a vector $\eta$ onto the tangent plane $S$, and $(\boldsymbol I-\boldsymbol P)$ can project any vector $\eta$ onto $S_T$. 

% The orthonormality of $\boldsymbol J_{\psi,x}$ implies that $\lVert
% \boldsymbol f_C
% \lVert
% =
% \lVert
% \boldsymbol \lambda
% \lVert$.
% %
% %
%
% Furthermore,  we derive the constraint Jacobian matrix $\boldsymbol J_{\psi}(\boldsymbol q,\theta)$ in the joint space as:
% \begin{equation}
% \boldsymbol J_{\psi}(\boldsymbol q,\theta) =
% \frac{\partial \boldsymbol \psi(\boldsymbol q,\theta)}
% {\partial {\boldsymbol q}}
% =
% \frac{\partial \boldsymbol \psi(\boldsymbol x,\theta)}
% {\partial {\boldsymbol x}}
% %
% %
% \frac{\partial \boldsymbol x}{\partial {\boldsymbol q}}
% =\boldsymbol J_{\psi,x}(\boldsymbol x,\theta)
% \boldsymbol J_C({\boldsymbol q})
% \right
% \end{equation}
% hence,  
% $\boldsymbol J^T_{\psi}(\boldsymbol q,\theta) \boldsymbol \lambda 
% =
% \boldsymbol J^T_C({\boldsymbol q})
% \boldsymbol J^T_{\psi,x}(\boldsymbol x,\theta)
% \boldsymbol \lambda 
% =
% \boldsymbol J^T_C({\boldsymbol q})
% \boldsymbol f_C.
% $
%
Furthermore, the actual joint toque ${\boldsymbol {\hat \tau}}_{\text{ext}}$ corresponds to the contact force ${\boldsymbol {\hat f}}_{\text{ext}}$ can be estimated by the general momentum observer  with the measurement of $\{\boldsymbol q, {\boldsymbol {\dot q}}, \boldsymbol \tau
\}$.
\begin{equation}
{\boldsymbol {\hat \tau}}_{\text{ext}}
=
F_{GM}
(\boldsymbol q, {\boldsymbol {\dot q}}, \boldsymbol \tau)
=
\boldsymbol{J}^T_e
{\boldsymbol {\hat f}}_{\text{ext}}
% \right
\end{equation}
where $F_{GM}$ is the general momentum observer algorithm function~\cite{focchi2020heuristic,bledt2018contact}. So ${\boldsymbol {\hat f}}_{\text{ext}}$ can be obtained by the inverse solution of the above equation. Thus, $\boldsymbol {\hat f}_C$ can be derived. %
In fact, the estimated $\boldsymbol {\hat f}_C$ usually contains a frictional force component. So the estimated normal vector
\begin{equation}\
   \boldsymbol {\hat n}_z={\boldsymbol {\hat f}_{C,z}}/{\lVert \boldsymbol {\hat f}_{C,z} \lVert}
\end{equation}
% What is f_r???????????????
and
\begin{equation}\
{\boldsymbol {\hat f}_{C,z}}={\boldsymbol {\hat f}_C-({\boldsymbol {\hat f}^T_C}{\boldsymbol {\hat n}_x}}){\boldsymbol {\hat n}_x}
\end{equation}

Then, $\boldsymbol {\hat n}_y$ can be derived as
\begin{equation}\
\boldsymbol {\hat n}_y=\boldsymbol {\hat n}_z 
\times 
\boldsymbol {\hat n}_x
\end{equation}

\subsection{The Model-free 
Compensation in WBC}
% 机器人本体由于关节摩擦，通信时延迟，机械死区等因素会影响任务跟踪性能。因此本文采用无模型的方法来自适应补偿相应的非线性项。
 We use a model-free approach to  derive   $\boldsymbol{\tau}_f$.
%
%
% Here we generalize algorithm in $A$ to the task space and combine it with WBC so that the robot can adaptively compensate for the tracking error caused by joint friction. 
%
% Furthermore, the joint friction torque can be expressed as
\begin{equation}
\boldsymbol{\tau}_f
=
\boldsymbol{J}^T
\boldsymbol \sigma(\boldsymbol{\dot {x}}(t),\boldsymbol u(t))
{\boldsymbol { F}}_f(t)
% \right
\end{equation}
where 
% $\boldsymbol{J}$ is the Jacobian matrix.
%
${\boldsymbol { F}}_f(t)$ is the time-varying friction value.
$\boldsymbol u(t)=\boldsymbol{\ddot x}_{\text{des}}$.
$\boldsymbol \sigma$ is the signum friction, and it  affects the direction of friction force in Eq.(18). its diagonal elements $\sigma_i$ is shown as
\begin{equation}
\label{eq:uiOptimal}
\sigma_i=
\left\{
\begin{array}{lcl}
1 ~~~\mathrm{if}~\dot x_i > 0~\mathrm{or}~ (\dot x_i=0~\mathrm{and}~u_i(t)>0)\\
-1 ~\mathrm{if}~\dot x_i < 0~ \mathrm{or}~ (\dot x_i=0~\mathrm{and}~u_i(t)<0)\\
0 ~~~\mathrm{if}~\dot x_i = 0~\mathrm{and}~u_i(t)=0 \\
\end{array} 
\right.
\end{equation}
It can be seen that the activation function is not only affected by $\boldsymbol {\dot x}$, but also affected by  $\boldsymbol u$ at zero speed. This can effectively avoid the defects of frictionless compensation corresponding to the zero-speed transition point of the trajectory.
Besides, the the update speed of differential value $ {\boldsymbol {\dot {  F}}}_f(t)$ is effected by  
$\boldsymbol \sigma$.
\begin{equation}
 {\boldsymbol {\dot {  F}}}_f(t)
=
\boldsymbol k_P \boldsymbol{\sigma}( {\boldsymbol {\dot e}}
+
{\boldsymbol k}_\lambda { \boldsymbol {e}})
% \right
\end{equation}
where $\boldsymbol k_P$ and $\boldsymbol k_\lambda$ are the tuning parameters.
$\boldsymbol {e}=\boldsymbol x_{\text{des}}(t)-\boldsymbol x(t)$ denotes the position error.
${\boldsymbol {{  F}}}_f(t)$ can be obtained by numerical integration of ${\boldsymbol {\dot {  F}}}_f(t)$.
\section{Experiments}
\label{sec:V}

\begin{figure*}[t]
\centering
\subfigure[\footnotesize Flat ground.]{
\begin{minipage}[t]{0.24\linewidth}
\centering
\includegraphics[width=1.5in]{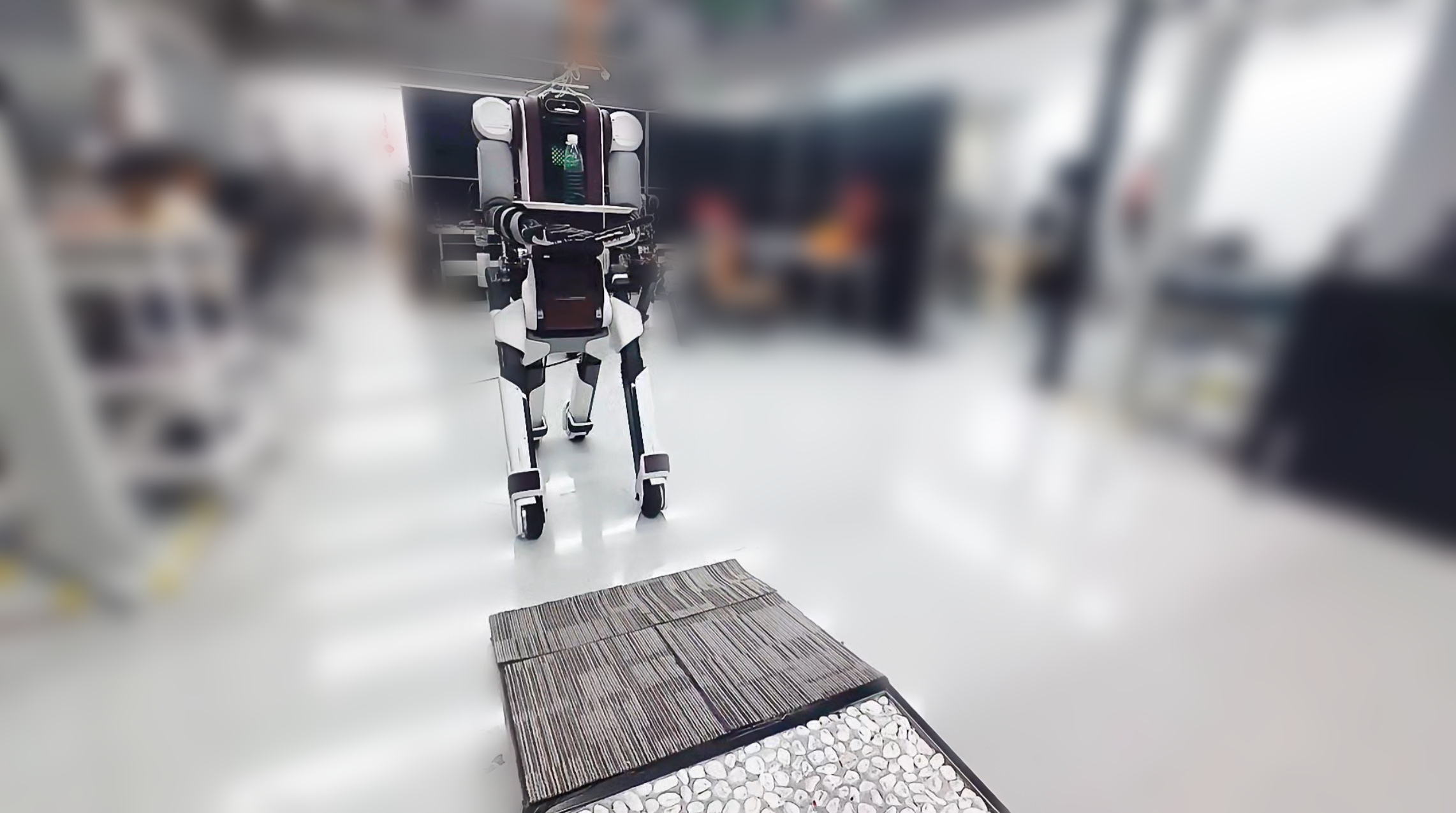}
%\caption{fig1}
\label{fig:de case2}
\end{minipage}%
}%
\subfigure[\footnotesize Uphill.]{
\begin{minipage}[t]{0.24\linewidth}
\centering
\includegraphics[width=1.5in]{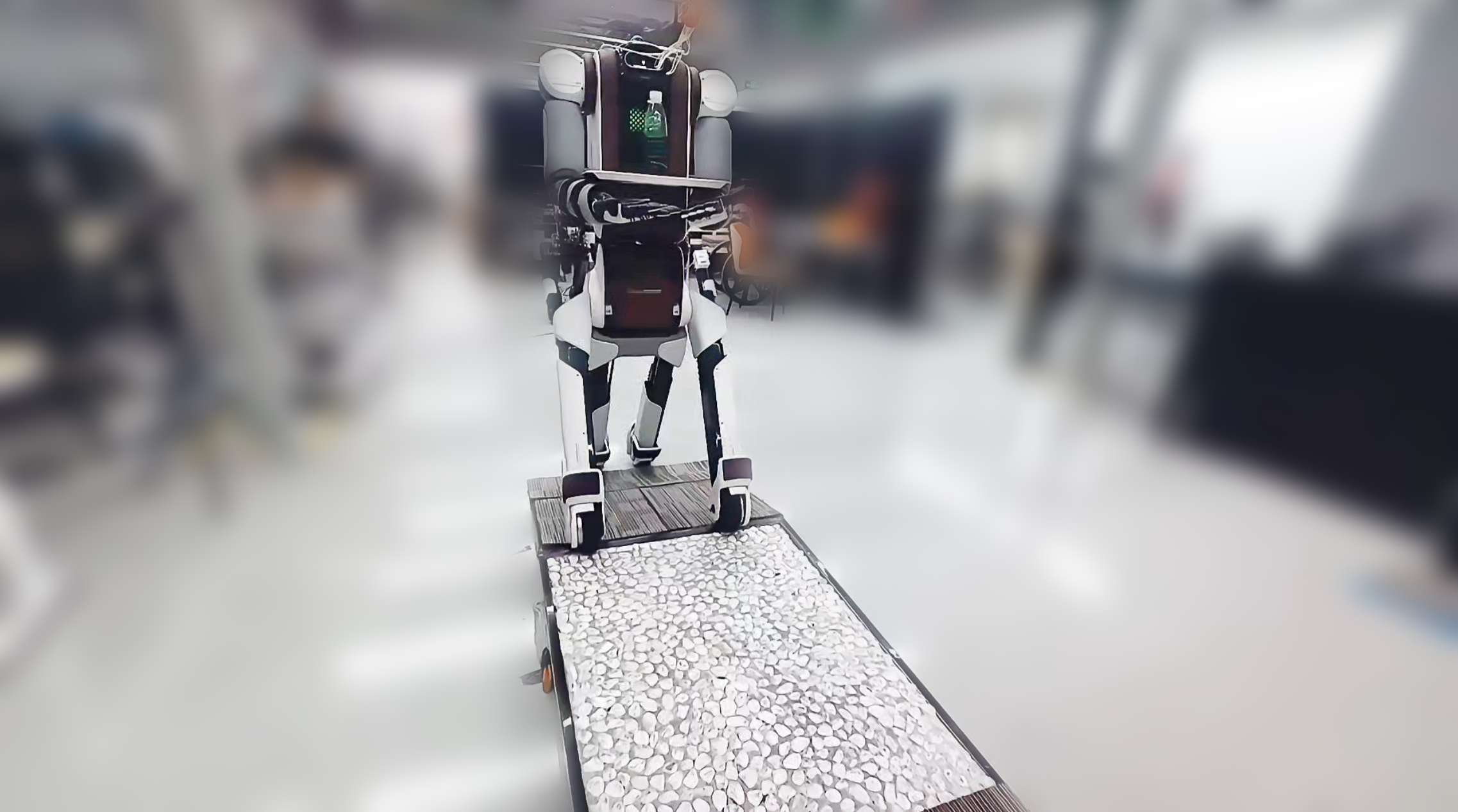}
%\caption{fig1}
\label{fig:de case2}
\end{minipage}%
}%
\subfigure[\footnotesize Cobblestone ground.]{
\begin{minipage}[t]{0.24\linewidth}
\centering
\includegraphics[width=1.5in]{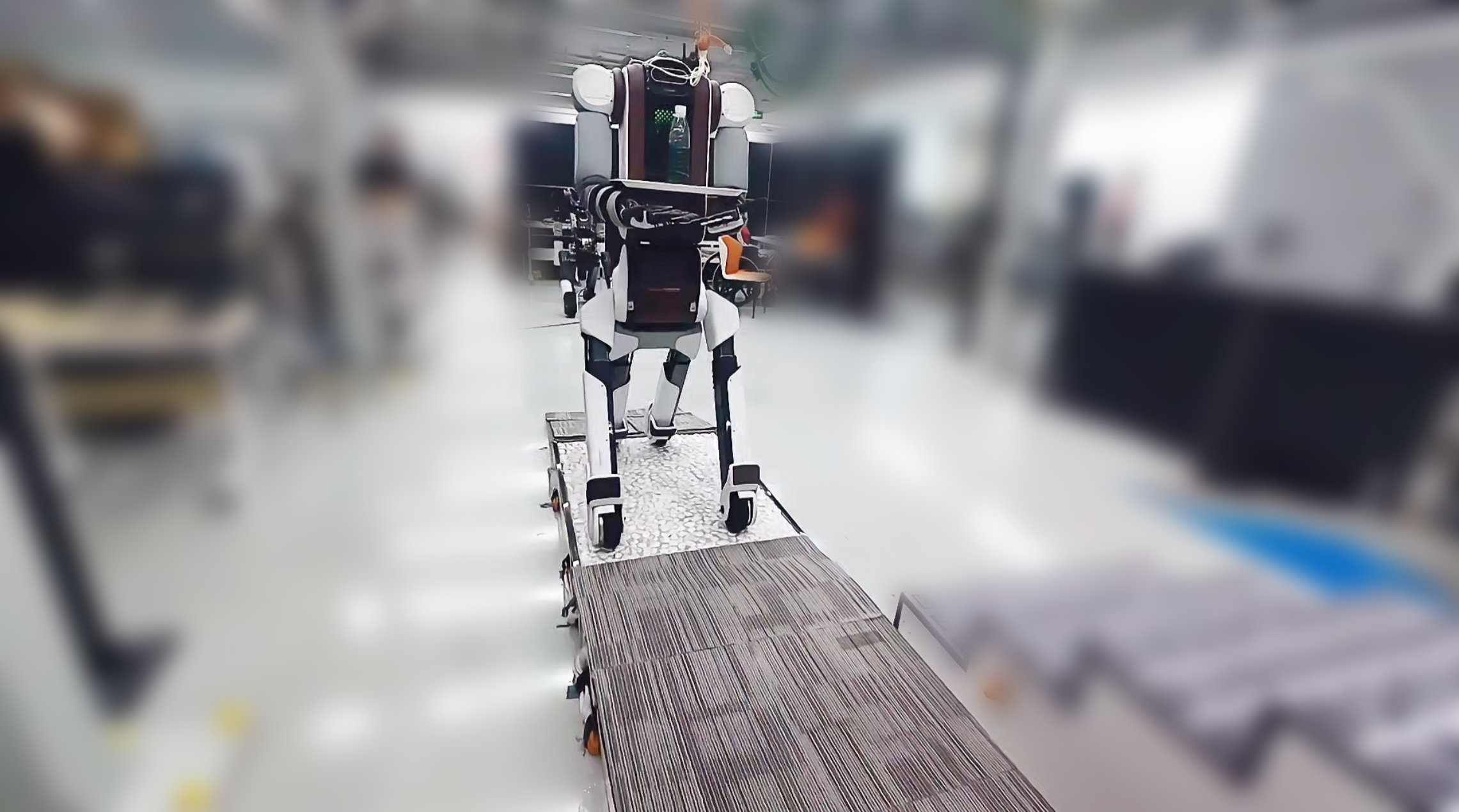}
%\caption{fig1}
\label{fig:de case2}
\end{minipage}%
}%
\subfigure[\footnotesize Downhill.]{
\begin{minipage}[t]{0.24\linewidth}
\centering
\includegraphics[width=1.5in]{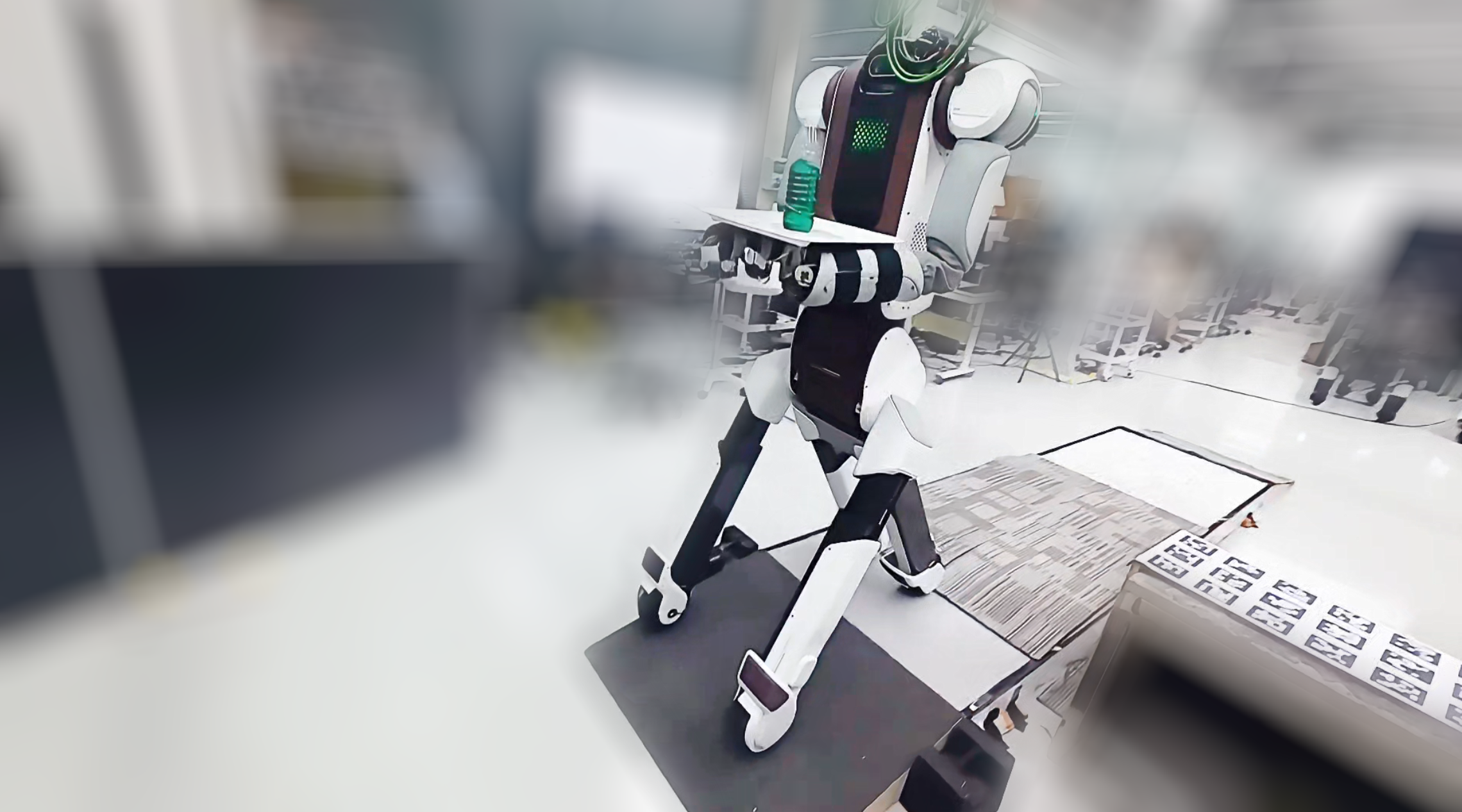}
%\caption{fig1}
\label{fig:de case2}
\end{minipage}%
}%
\vspace{-0.3cm}

\centering
\caption{\footnotesize Task A. Whole body compliance control experiment in the terrain 1(uphill-cobblestone-downhill road).}
\vspace{-0.2cm}
\label{taskA}
\end{figure*}

\begin{figure*}[t]
\centering
\subfigure[\footnotesize Flat ground.]{
\begin{minipage}[t]{0.24\linewidth}
\centering
\includegraphics[width=1.5in]{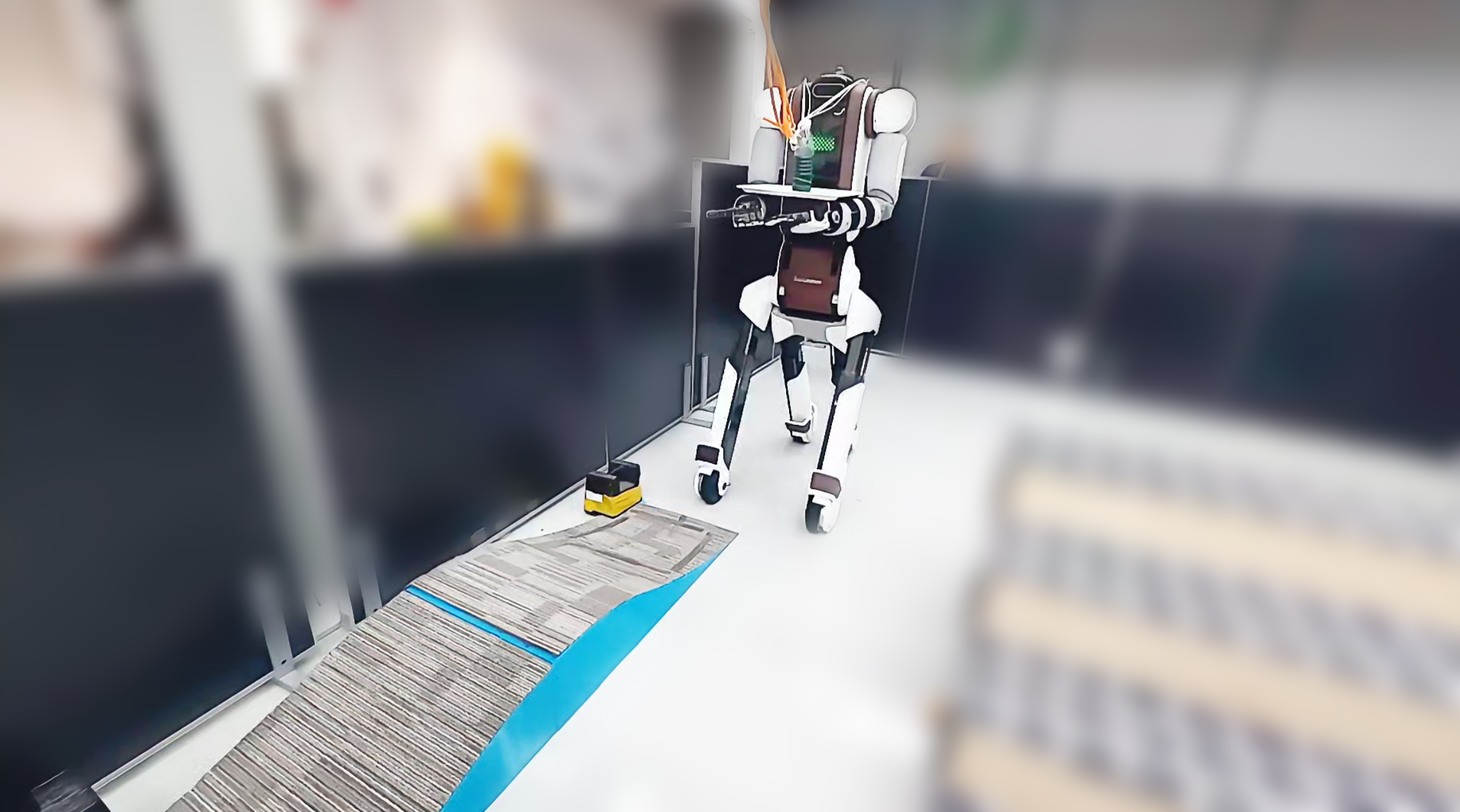}
%\caption{fig1}
\label{fig:de case2}
\end{minipage}%
}%
\subfigure[\footnotesize Wave slope phase 1.]{
\begin{minipage}[t]{0.24\linewidth}
\centering
\includegraphics[width=1.5in]{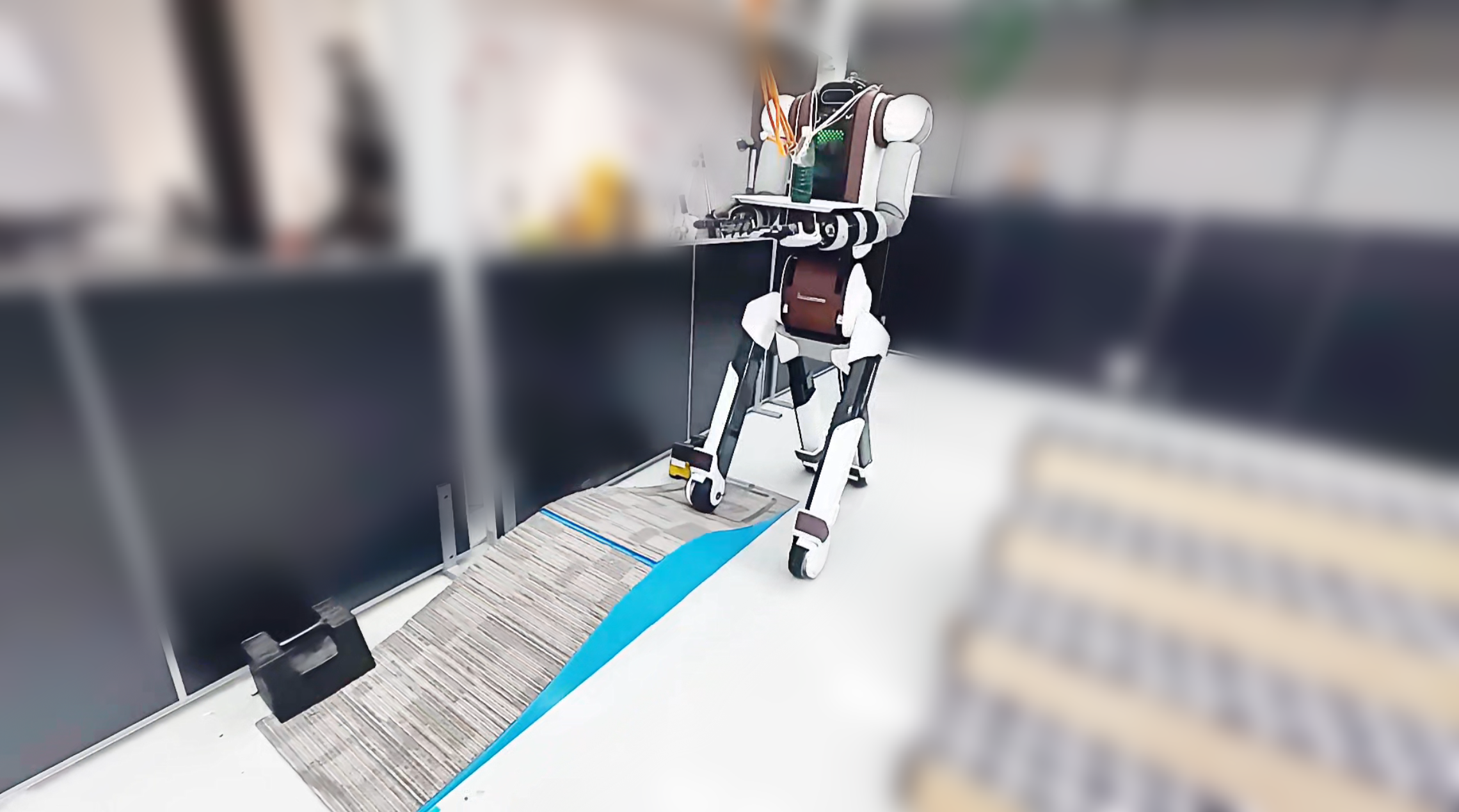}
%\caption{fig1}
\label{fig:de case2}
\end{minipage}%
}%
\subfigure[\footnotesize Wave slope phase 2.]{
\begin{minipage}[t]{0.24\linewidth}
\centering
\includegraphics[width=1.5in]{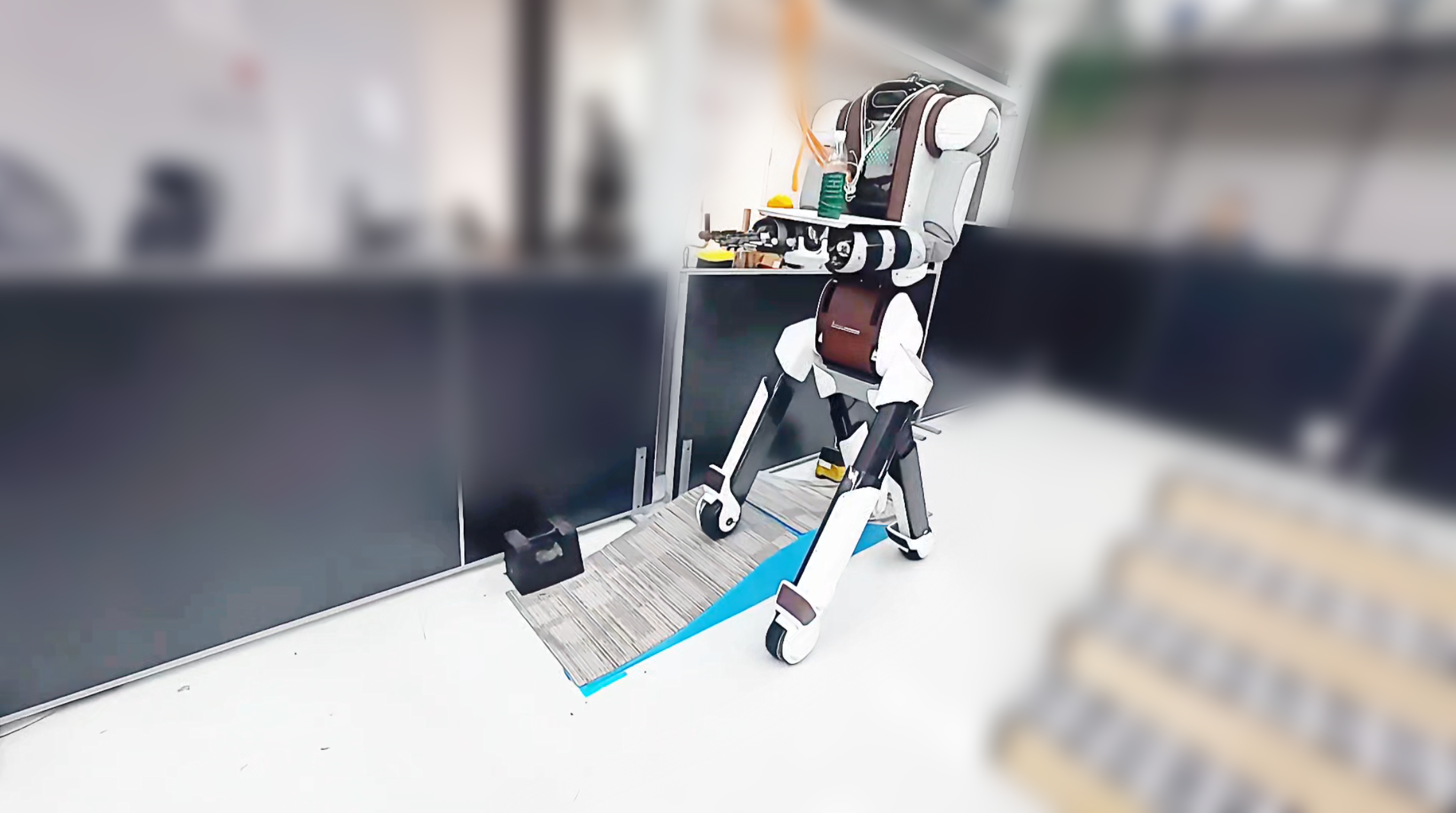}
%\caption{fig1}
\label{fig:de case2}
\end{minipage}%
}%
\subfigure[\footnotesize Wave slope phase 3.]{
\begin{minipage}[t]{0.24\linewidth}
\centering
\includegraphics[width=1.5in]{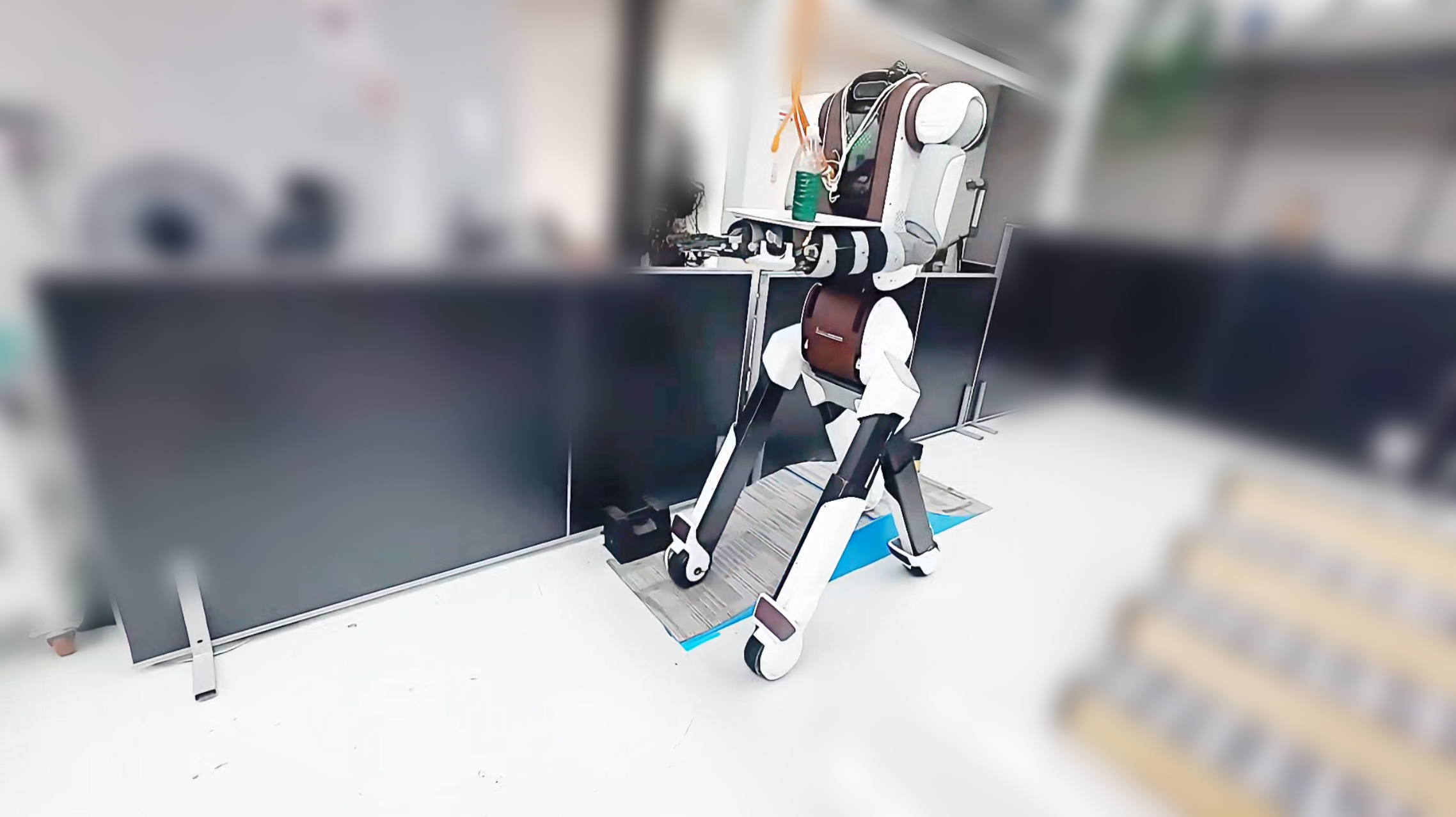}
%\caption{fig1}
\label{fig:de case2}
\end{minipage}%
}%
\vspace{-0.3cm}

\centering
\caption{\footnotesize Task B. Whole body compliance control experiment in the terrain 2(wave slope road).(phase $1$ means the front wheel goes uphill, phase $2$ means the front wheel goes downhill and the rear wheel goes uphill, and phase $3$ means the rear wheel goes downhill.)}
\vspace{-0.2cm}
\label{taskB}
\end{figure*}

% \subsection{The Experiment in Humanoid Robot}

The WBC setup described in this article is implemented in C++. The software uses the Eigen3 library and the qpOASES quadratic progrmming library. 
The proposed controller was running on a 8-cores Intel Core i7-32G RAM computer with 500
$Hz$ control loop.

% 实验场景的设定是，机器人通过平底，上下坡，鹅暖石地面，波浪坡等地行，在此同时，双臂端着托盘，托盘上放水，整个过程中尽量保证上半身稳定，水的晃动小。 
The experimental scene is set up in such a way that the robot walks on flat ground, uphill and downhill, cobblestone ground, wave slopes, etc. 
%
% 考虑腿部的最大伸缩量35cm，实验场地的最大高度差设定为20cm。
Considering the maximum extension and contraction of the legs is 35~$\mathrm{cm}$, the maximum height difference of the experimental site is set to 20~$\mathrm{cm}$. The friction coefficient of the robot's tires with various terrains is between 0.06 and 0.12.

Here we set the friction coefficient in the algorithm to 0.1, and the uncertainty of the friction coefficient is guaranteed by the robustness of the WBC algorithm and the slip detection algorithm.
During the experiment, the robot did not know the specific information about the slope and friction of the terrain.
%
% At the same time, both arms hold a tray with water on it.
% 为了表现不确定地形信息下机器人操作物体的稳定性，我么选择双臂端盘子，盘子上放瓶水。盘子的尺寸是30x20，水的重量是0.6kg。盘子和瓶子的摩擦系数为0.5，双臂和盘子的摩擦系数是1.2。
In order to demonstrate the stability of the robot manipulating objects under uncertain terrain information, we choose to hold a plate with both arms and place a bottle of water on the plate. The size of the plate is 38.5x28.5~$\mathrm{cm}$, and the weight of the water is 0.6~$\mathrm{kg}$. The friction coefficient between the plate and the bottle is 0.5, and the friction coefficient between the arms and the plate is 1.2.
%
%
% 整个过程中，我们预设了最大前进速度1，最大转弯速度0.5。
%
Throughout the process, we preset the maximum forward speed to 0.8~$\mathrm{m/s}$ and the maximum turning speed to 0.5~$\mathrm{rad/s}$. The actual navigation instructions are calculated in real time by the ROS navigation module based on the SLAM perception information.
%
%
% During the whole process, the control algorithm needs to ensure that the upper body is stable and the water shakes little.
%
% 值得一提的是，机器人除了各个关节收到关节摩擦之外，轮子处还有地面和轮子之间的摩擦，以主动轮在前为例，主动轮的转向控制还受到被动轮子的侧向摩擦力产生的力矩。

% %
% It is worth mentioning that in addition to the joint friction of each joint of the robot, there is also friction between the ground and the wheels. Taking the driving wheel in the front as an example, the steering control of the driving wheel is also affected by the torque generated by the lateral friction force of the passive wheel.

%
We set the parameters in the ICC as: 
$M=80$, $D_b=600$, $K_b=100000$, $m_L=0.9$, $D_{L,\text{min}}=20$, $D_{L,\text{max}}=200$, $K_L=350$,
$m_R=0.9$, $D_{L,\text{mix}}=20$, $D_{L,\text{max}}=200$, $K_R=350$.
And we also set $D_{L}=100$, $D_{R}=100$ for comparison in ablation experiments.. 
The configuration-related task parameters of the robot task in the WBC are set as: the proportional and derivative parameters of the robot height task are 1000 and 20 respectively. The proportional and derivative parameters of the robot wheel-Centroid task are 800 and 10 respectively. The proportional and derivative parameters of the robot base rotation task are 300 and 15 respectively.
These sets of parameters will directly determine the overall configuration of the robot.

During the entire experiment, the robot did not know the accurate information of the terrain. The upper body of the robot remained vertical based on the perception of the IMU, and the overall configuration of the robot remained unchanged.
The experimental screenshots are shown in the Fig.~\ref{taskA} and Fig.~\ref{taskB}. 
The experimental terrain shape and terrain slope curves are shown in the Fig. 8. Terrain-1 mainly includes uphill, cobblestone and downhill. Terrain-2 mainly consists of wave slope.
It can be seen that, through the algorithm given in Sec.V.B, the robot successfully estimates the terrain information and introduces it into the WBC solution.
In the Fig.8, the maximum slope of the terrain is 0.22~$\mathrm{rad}$.
%
% 从实验截图可以看出来，机器人在整个过程中维持高度不变，除非无法通过地形时候才会升高高度（高度升高，腿部伸长，进而地形适应的裕量增加）。机器人在上坡时候前腿会缩短以适应高度不变，下坡时候后腿缩短以适应高度。并且在过波浪坡的时候腿部会有明显的交替缩短伸长的变化。
%
In the experiment, the robot maintains a constant height throughout the entire process, and only increases its height when it cannot pass the terrain. 
In the Fig.~\ref{legcurve}, when going uphill, the robot's front legs shorten to adapt to the constant height, and when going downhill, the front legs stretch to adapt to the height. And when passing a wave slope, the legs will have obvious changes in alternating shortening and lengthening.

\begin{figure}[htbp]
\captionsetup{font={footnotesize}}
\centering %
\includegraphics[width=0.5\textwidth]{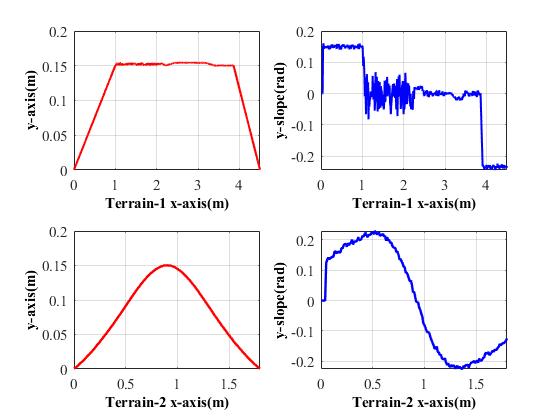}
\caption{Curves of terrain shape (shown in up left and down left) and terrain slope (shown in up right and down right).}
\vspace{-0.4cm}
\label{terrain_shape}
\end{figure}

\begin{figure}[htbp]
\captionsetup{font={footnotesize}}
\centering %
\includegraphics[width=0.5\textwidth]{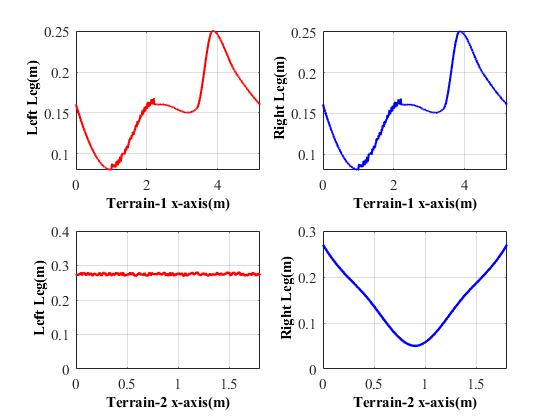}
\caption{Extension curves of the front legs in the terrain 1 and 2.}
\vspace{-0.4cm}
\label{legcurve}
\end{figure}

\begin{figure}[htbp]
\captionsetup{font={footnotesize}}
\centering %
\includegraphics[width=0.5\textwidth]{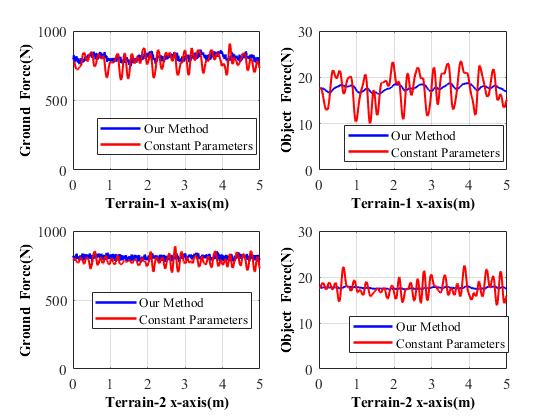}
\caption{Curves of ground contact force(shown in up left and down left) and object contact forces(shown in up right and down right).}
\vspace{-0.4cm}
\label{groundcontactforce}
\end{figure}

% 从实验结果可以看出，机器人在整个过程中可以在穿越不同地形的同时保持上半身运输水的稳定性。从实验曲线可以看出来，在变阻尼控制器的作用下，机器人的上半身的接触力较为平顺，被控制在合理的范围（力矩的波动在5N范围内）。
%
The experimental curves shown in Fig.~\ref{groundcontactforce} shows that under the action of the variable damping controller, the contact force of the robot's upper body is relatively smooth and controlled within a reasonable range (the force fluctuation is within 18~$\mathrm{N}$). The results show that the robot can maintain the stability of the upper body transporting water while crossing different terrains.
Through the algorithm in this article, the force of the robot's upper body can be effectively controlled within a reasonable range to avoid the impact of terrain changes on the robot.

\section{Conclusions}
\label{sec:VI}

This article proposes an impedance coordinative control for our newly developed wheel-legged humanoid robot in order to cope with locomotion tasks. Such impedance coordinative control also allows the robot to maintain upper-body passive stability when carrying objects, even in challenging terrain environments.
% This article proposes a method that combines impedance control and WBC, and adopts upper body variable damping control, which can effectively ensure the force control stability of the robot in locomotion manipulation.
In addition, a whole-body control framework for robots that includes terrain information update and model-free compensation is proposed. The update of terrain information can timely correct the tangential and normal information of the robot's contact point with the ground, and then update the range of the friction cone constraint. 
%
%
%
%
%
% 。
% A whole-body compliant control experiment was carried out using a quadruped humanoid robot developed by ourselves.
Experimental results verify that the proposed algorithms effectively support the robot in adapting to various terrains while ensuring upper-body passive stability.

% 我们未来的工作将会考虑上半身的6D的变阻尼阻抗控制，以使得机器人在适应地形的同时可以在六个方向均保证上半身的稳定性。
% Our future work will consider: 
%
% 1) 6D variable damping impedance control of the upper body,
 % multi-modal control of X-man, such as four-wheel-two-wheel mode, wheel-foot hybrid mode, etc.

% \section{Future Works}

%%%%%%%%%%%%%%%%%%%%%%%%%%%%
\bibliographystyle{IEEEtran}
\normalem
\bibliography{ref}

    \end{document}